\documentclass{article} % For LaTeX2e
\usepackage{iclr2025_conference,times}

\usepackage{amsmath,amsfonts,bm}

\def\eqref#1{equation~\ref{#1}}
\def\1{\bm{1}}

\DeclareMathAlphabet{\mathsfit}{\encodingdefault}{\sfdefault}{m}{sl}
\SetMathAlphabet{\mathsfit}{bold}{\encodingdefault}{\sfdefault}{bx}{n}

\usepackage{hyperref}
\usepackage{url}
\usepackage{booktabs}       % professional-quality tables
\usepackage{amsfonts}       % blackboard math symbols
\usepackage{nicefrac}       % compact symbols for 1/2, etc.
\usepackage{microtype}      % microtypography
\usepackage{xcolor,colortbl}% colors
\usepackage{graphicx}
\usepackage[ruled,linesnumbered]{algorithm2e}
\usepackage{algpseudocode}
\usepackage[english]{babel}
\usepackage{amsthm}

\usepackage{comment}
\usepackage{bm}
\usepackage{amssymb}
\usepackage{amsmath}
\usepackage{bbm}
\usepackage{xspace}
\usepackage{amsthm}
\usepackage{mathtools}
\usepackage{dsfont}
\usepackage{wrapfig}

\usepackage{multirow}

\usepackage{parskip}
\definecolor{Gray}{gray}{0.9}
\definecolor{LightCyan}{rgb}{0.88,1,1}

\newcolumntype{a}{>{\columncolor{Gray}}c}
\newcolumntype{b}{>{\columncolor{white}}c}

\title{Efficient Model Editing with Task-Localized Sparse Fine-Tuning}

\author{Leonardo Iurada$^{1,*}$, Marco Ciccone$^2$, Tatiana Tommasi$^1$ \\
$^1$Politecnico di Torino, Italy \hspace{5mm} $^2$Vector Institute, Toronto, Ontario, Canada \\
$^*$Correspondance to: \texttt{leonardo.iurada@polito.it} \\
}

\makeatletter
\DeclareRobustCommand\onedot{\futurelet\@let@token\@onedot}
\def\@onedot{\ifx\@let@token.\else.\null\fi\xspace}

\def\eg{\emph{e.g}\onedot} 
\def\ie{\emph{i.e}\onedot}

\def\iid{i.i.d\onedot} 

\makeatother

\definecolor{difcolor}{RGB}{204, 0, 204}

\newcommand{\up}[1]{\scriptsize{[+\textcolor[HTML]{00BB00}{#1}]}}
\newcommand{\dw}[1]{\scriptsize{[-\textcolor[HTML]{DD0000}{#1}]}}
\newcommand{\na}[1]{\scriptsize{[ \textcolor[HTML]{000000}{#1}]}}

\usepackage{makecell}

\usepackage{enumitem}

\newcommand{\ul}[1]{\underline{#1}}
\newcommand{\tb}[1]{\textbf{#1}}

\newcommand{\longname}{\textit{Task-Localized Sparse Fine-Tuning}\xspace}
\newcommand{\shortname}{\texttt{TaLoS}\xspace}

\iclrfinalcopy % Uncomment for camera-ready version, but NOT for submission.
\begin{document}

\maketitle

\begin{abstract}
    Task arithmetic has emerged as a promising approach for editing models by representing task-specific knowledge as composable task vectors. However, existing methods rely on network linearization to derive task vectors, leading to computational bottlenecks during training and inference. 
    Moreover, linearization alone does not ensure weight disentanglement, the key property that enables conflict-free composition of task vectors.
    To address this, we propose \shortname which allows to build 
    sparse task vectors with minimal interference without requiring explicit linearization and sharing information across tasks. We find that pre-trained models %transformers 
    contain a subset of parameters with consistently low gradient sensitivity across tasks, and that sparsely updating only these parameters allows for promoting weight disentanglement during fine-tuning.
    Our experiments prove that \shortname improves training and inference efficiency while outperforming current methods in task addition and negation. By enabling modular parameter editing, our approach fosters practical deployment of adaptable foundation models in real-world applications\footnote{Code available at: \url{https://github.com/iurada/talos-task-arithmetic}}.
\end{abstract}

\section{Introduction}

Large pre-trained models \citep{radford2021clip, 2020t5, brown2020gpt3} have become the cornerstone of modern machine learning, showcasing impressive capabilities across a broad spectrum of tasks. Currently, their development is confined to a few computationally and financially well-resourced research groups, but once publicly released they provide a wealth of reusable knowledge that greatly benefits downstream applications. 
Indeed, fine-tuning large models to achieve optimal performance on specialized tasks or to align with user preferences is becoming an increasingly democratized practice, thanks to efficient methods enabling model customization on affordable consumer GPUs. Parameter-efficient fine-tuning (PEFT) \citep{hu2022lora, liu2022few, liu24dora}, sparsity \citep{ansell-etal-2022-composable, ansell2024scaling}, and quantization \citep{dettmers2024qlora} are some of the techniques that fueled the growth of a rich ecosystem of task-specific models.
They are, in turn, readily shared on open platforms \citep{pfeiffer2020AdapterHub, poth-etal-2023-adapters}  fostering collaborative knowledge building by enabling users to adapt and integrate specialized modules \citep{raffel2023building}.

In this context, \emph{task arithmetic} \citep{ilharco2023editing} has emerged as a promising framework for scalable and cost-effective model editing. It encodes task-specific knowledge using \emph{task vectors}, derived by fine-tuning a pre-trained model and subtracting its original weights from the fine-tuned ones. Task vectors can be combined through addition and subtraction to enhance specific tasks, suppress undesired behaviors, or merge functionalities. However, when task vectors are independently fine-tuned in decentralized collaborative settings, task interference becomes a significant concern \citep{yadav2023ties, wang2024consensus}, as adding or removing a functionality disrupts previously acquired knowledge. Task interference occurs when fine-tuning modifies parameters that are critical to other tasks, resulting in unintended behavioral shifts. To prevent this, data from disjoint regions in the input space (representing different tasks) should affect only their corresponding regions in the activation space. \citet{Tangent_task_arith_2023} formalized this concept as \textit{weight disentanglement}. Their research showed that this property is an emergent feature of pre-training, which makes foundation models inherently suited for task arithmetic. The key question therefore becomes: \emph{how can fine-tuning preserve weight disentanglement?}

Explicitly linearizing the model during fine-tuning has been a promising direction to maintain weight disentanglement, albeit with increased computational overhead \citep{Tangent_task_arith_2023}.
In this work, we first show that model linearization alone is not sufficient, as its task functions can still activate for arbitrary inputs. Instead, we propose a set of \emph{function localization} constraints to exactly implement the weight disentanglement property on linearized networks. Then, we introduce a novel \textit{sparse fine-tuning} approach that implements such constraints while avoiding the need for explicit model linearization. The proposed method strategically updates a subset of model parameters, simultaneously promoting linearized behavior and enforcing function localization. Extensive empirical analyses and theoretical justifications demonstrate that our approach \textit{effectively promotes weight disentanglement}, ensuring compatibility between task vectors without the need for sharing information between users and tasks. This enables efficient and robust model editing through the simple addition and subtraction of sparse task vectors, facilitating decentralized collaborative strategies.

\textbf{We can summarize our main contributions as follows.}
\begin{itemize}[leftmargin=*]
    \item We advance the field of task arithmetic by deriving a novel set of function localization constraints that provide exact guarantees of weight disentanglement on linearized networks.

    \item We empirically observed that the least sensitive parameters in transformer-based architectures pre-trained on large-scale datasets can be consistently identified regardless of the task. We exploit this regularity to satisfy the localization constraints under strict individual training assumptions.
    
    \item We introduce \longname (\shortname) that enables task arithmetic by jointly implementing the localization constraints and inducing a linear regime during fine-tuning, without incurring in the overheads of explicit network linearization.
\end{itemize}
Overall, our work addresses a critical gap in task arithmetic, providing a more complete and practical framework for parameter-space model editing, targeting real-world applications.
\vspace{-2.5mm}

\section{Related Works}

\textbf{Sparsity \& Parameter-Efficient Fine-Tuning.} 
Sparsity has emerged as a fundamental concept in efficient deep learning, manifesting in both training and adaptation methodologies. Sparse fine-tuning strategies \citep{guo-etal-2021-parameter, xu-etal-2021-raise} improve training efficiency by selectively updating subsets of model parameters. These approaches often leverage the Fisher information matrix \citep{fisher1922mathematical, amari1996neural} to identify important weights for updating \citep{sung2021training, ben-zaken-etal-2022-bitfit} or, conversely, focus on fine-tuning only the least important parameters to minimize disruption of the original model's knowledge \citep{liao-etal-2023-parameter, ansell2024scaling}. Sparse masking techniques \citep{wortsman2020supermasks, mallya2018piggyback, mallya2018packnet, havasi2020training} further exploit this principle by employing subnetworks for continual and multi-task learning.
Parameter-efficient fine-tuning (PEFT) represents another approach to adaptation with minimal parameter updates. Popular PEFT methods include adapter layers \citep{pmlr-v97-houlsby19a}, prefix tuning \citep{prefix2021}, and low-rank adaptation (LoRA, \citep{hu2022lora}). LoRA in particular approximates model updates through rank decomposition matrices while keeping pre-trained weights frozen. In a complementary direction, \citet{ansell-etal-2022-composable, panda2024lottery} investigate sparse weight addition as a flexible approach to model composition.
These sparse adaptation techniques connect to the broader field of model pruning, which has traditionally been applied post-training for efficient storage and inference \citep{blalock2020state}. The Lottery Tickets Hypothesis \citep{frankle2018lottery} expanded this idea by demonstrating that sparse subnetworks identified at initialization can, when trained, match the performance of the original dense model while significantly reducing computational costs.

\textbf{Model Merging.} 
The goal of model merging is to combine multiple task-specific models into a single multi-task model without performing additional training. This requires merging techniques that prevent negative interferences among separately learned parameters. While simple parameter averaging can be effective, particularly when fine-tuned models share the same initialization \citep{wortsman2022model, rame2023model}, it does not always yield optimal results. As a result, existing approaches explored tailored re-weighting schemes, though these often come with high computational costs. 
RegMean \citep{jin2023dataless} solves a local linear regression problem for each individual linear layer in the model that requires transmitting extra data statistics of the same size as the model and additional inference steps. 
Fisher Merging \citep{mergingfisher_2024} exploits the Fisher Information Matrix. This method, however, requires computing gradients, resulting in high memory costs.
A recent approach exploits extra unlabeled data to learn the model merging weights \citep{AdaMerging_ICLR_2024}. 

\textbf{Task Arithmetic.} Task arithmetic \citep{ilharco2023editing} was introduced as a paradigm for editing models based on arithmetic operations over \emph{task vectors} obtained by fine-tuning a base pre-trained model and then subtracting the pre-trained weights from the fine-tuned ones. This concept has also been used in model merging, with methods that prepare task vectors before adding them together to produce a single multi-task model. Recent examples of this strategy are 
TIES-Merging \citep{yadav2023ties} which resolves parameter overlap and sign conflicts after merging using heuristics, and TALL Masks / Consensus \citep{wang2024consensus} that deactivates irrelevant parameters through binary masking. Other approaches sparsify task vectors by randomly dropping and rescaling parameters \citep{dareSupermario} or masking weight outliers \citep{davari2023model}. 
However, task arithmetic goes beyond model merging as it aims at \emph{adding to} or \emph{deleting} knowledge and capabilities \emph{from} a model in a modular and efficient manner. 
Its effectiveness relies on weight disentanglement, a property emerging during pre-training, as shown by \citet{Tangent_task_arith_2023}. They proposed to preserve weight disentanglement by fine-tuning in the tangent space via full model linearization with high computational costs. 
To improve efficiency, \citet{tang2023parameter} proposed to use linearized low-rank adapters in the attention modules during fine-tuning. 
Still, linearization alone does not guarantee task localization, potentially letting weight disentanglement decrease during fine-tuning. 

Our work fits within task arithmetic as a PEFT approach to construct \textit{sparse task vectors}. By leveraging strategies from pruning and sparse fine-tuning, we introduce a parameter update criterion that induces a linearized regime without explicit linearization and ensures functional task localization.

\section{Background}
\label{sec:background}

Consider a neural network $f$ with parameters $\bm{\theta}\in \mathbb{R}^m$, pre-trained on a mixture of tasks $\mathcal{P}$ to obtain parameters $\bm{\theta}_0$. We are interested in fine-tuning the pre-trained model $f(\cdot, \bm{\theta}_0)$ on a set of $T$ distinct classification tasks, with associated non-intersecting task data support $\mathcal{D} = \{ \bigcup_{t=1}^T \mathcal{D}_t \} \subseteq \mathcal{D}_{\mathcal{P}}$ (\ie $\forall t,t'$ if $t \neq t'$ then $\mathcal{D}_t \cap \mathcal{D}_{t'} = \varnothing$).

In this setting, the core idea behind task arithmetic, introduced in \cite{ilharco2023editing}, is to represent the knowledge acquired for each task $t$ as a \textit{task vector} $\bm{\tau}_t=\bm{\theta}^\star_t - \bm{\theta}_0$, obtained by subtracting the initial parameters from the fine-tuned parameters. Intuitively, this vector captures the direction and magnitude of change in the model's weight space induced by learning task $t$. By manipulating tasks via task arithmetic operations we can effectively add, combine, or remove knowledge in the pre-trained model producing actual functional behaviors directly in the parameters space.

As formalized by \citet{Tangent_task_arith_2023}, a network $f$ is said to satisfy the task arithmetic property around $\bm{\theta}_0$ if it holds
\begin{equation}
    f \left(\bm{x}, \bm{\theta}_0 + \sum_{t=1}^T \alpha_t \bm{\tau}_t\right) = \begin{cases}
        f(\bm{x}, \bm{\theta}_0 + \alpha_t \bm{\tau}_t) &\bm{x} \in \mathcal{D}_t \\
        f(\bm{x}, \bm{\theta}_0) &\bm{x} \notin \bigcup_{t=1}^T \mathcal{D}_t
    \end{cases}
    \label{eq:1}
\end{equation}
with scaling factors $(\alpha_1, ..., \alpha_T) \in \mathcal{A} \subseteq \mathbb{R}^T$. This equation essentially states that adding a linear combination of task vectors to the initial parameters $\bm{\theta}_0$ is equivalent to selectively applying each task-specific modification to the model. 
In other words, the performance of the pre-trained model on different tasks can be modified independently if the task vector $\bm{\tau}_t$ does not modify the output of the model outside $\mathcal{D}_t$.

To fulfill the task arithmetic property, \citet{Tangent_task_arith_2023} states that the model $f$ must exhibit a form of \textit{weight disentanglement} with respect to the set of fine-tuning tasks, \textit{i.e.}, $f$ should behave as a composition of spatially localized components corresponding to functions that vanish outside the task's data support. 
In practice, Equation \ref{eq:1} can be re-written as 
\begin{align}
f \left(\bm{x}, \bm{\theta}_0 + \sum_{t=1}^T \alpha_t \bm{\tau}_t\right) & = f(\bm{x}, \bm{\theta}_0)\mathds{1}\left(\bm{x} \notin \bigcup_{t=1}^T \mathcal{D}_t \right) + \sum_{t=1}^T f(\bm{x}, \bm{\theta}_0 + \alpha_t \bm{\tau}_t)\mathds{1}(\bm{x} \in \mathcal{D}_t ) \\
& = g_0(\bm{x}) + \sum_{t=1}^Tg_t(\bm{x},\alpha_t\bm{\tau}_t)~.
\label{eq:3}
\end{align}
where $g_t(\bm{x}, \alpha_t \bm{\tau}_t) = \bm{0}$ for $\bm{x} \notin \mathcal{D}_t$ and $t=1,...,T$, and $g_0(\bm{x}) = 0$ for $\bm{x} \in \bigcup_{t=1}^T \mathcal{D}_t$, capturing the base behavior of the pre-trained model on inputs outside any of the task support.

%%%%%%%%%%%%%%%%%%%%%%%%

Previous works~\citep{tang2023parameter, Tangent_task_arith_2023} have sought to achieve task arithmetic by focusing on linearized neural networks~\citep{ortiz2021can}, as they explicitly constrain $f$ to be represented as a linear combination of functions.
Specifically, the linearization of $f$ can be achieved by its first-order Taylor expansion centered around $\bm{\theta}_0$:
\begin{equation}
    f(\bm{x}, \bm{\theta}_0 + \alpha_t\bm{\tau}_t) \approx f_\text{lin}(\bm{x},\bm{\theta}_0 + \alpha_t\bm{\tau}_t) = f(\bm{x}, \bm{\theta}_0) + \alpha_t\bm{\tau}_t^\top \nabla_{\bm{\theta}} f(\bm{x}, \bm{\theta}_0) ~. 
\end{equation}
The model $f_\text{lin}(\bm{x},\bm{\theta}_0 + \bm{\tau}_t)$ represents a linearized neural network. For this type of networks, when combining together multiple task vectors, it holds
\begin{equation}\label{eq:lin_net_addition}
    f_\text{lin}\left(\bm{x}, \bm{\theta}_0 + \sum_{t=1}^T \alpha_t \bm{\tau}_t\right) = f(\bm{x}, \bm{\theta}_0) + \sum_{t=1}^T \alpha_t \bm{\tau}_t^\top \nabla_{\bm{\theta}} f(\bm{x}, \bm{\theta}_0) ~. 
\end{equation}
While Equation \ref{eq:lin_net_addition} appears to closely resemble the weight disentanglement condition presented in Equation \ref{eq:3}, this similarity is superficial unless each term $\alpha_t \bm{\tau}_t^\top \nabla{\bm{\theta}} f(\bm{x}, \bm{\theta}_0)$ corresponds to a function that vanishes outside its task data support (\ie it is \emph{localized} within $\mathcal{D}_t$). In the following, we will demonstrate how to efficiently impose a condition of function localization.

%%%%%%%%%%%%%%%%%%%%%%%%

\section{Task-Localized Sparse Fine-Tuning}
\label{sec:method}

To formalize the condition of function localization for task arithmetic, we begin by revisiting the linear approximation of $f$ used in linearized fine-tuning. For Equation \ref{eq:lin_net_addition} to satisfy the weight disentanglement conditions in Equation \ref{eq:3}, we must ensure that each $t$-th task-specific function $\bm{\tau}_t^\top \nabla_{\bm{\theta}} f(\bm{x},\bm{\theta}_0)$ is active (non-zero) only for inputs within its corresponding task support, \ie, $\bm{x} \in \mathcal{D}_t$. This requirement can be expressed as a set of constraints:
\begin{equation}\label{eq:func_loc_constraint}
\forall \bm{x}\in \mathcal{D}_{t'\neq t},
~~~\bm{\tau}_t^\top \nabla_{\bm{\theta}} f(\bm{x},\bm{\theta}_0) = 0 ~.
\end{equation}
Satisfying these conditions ensures that updating the model's weights by training on task $t$ does not affect how the model processes data from other tasks, preventing interference between task vectors.

Directly implementing Equation \ref{eq:func_loc_constraint} poses a significant practical challenge. Enforcing the constraint $\forall \bm{x} \in \mathcal{D}_{t'}$  requires simultaneous access to data from all other tasks ($t' \neq t$) during fine-tuning on task~$t$. However, this is an impractical requirement in realistic settings where contributors optimize their model asynchronously on private, task-specific data.
To address this, we assume that during pre-training the model is exposed to a vast mixture of tasks, including some that are similar to the $T$ fine-tuning tasks under consideration. Consequently, we expect the gradients $\nabla_{\bm{\theta}} f(\cdot,\bm{\theta}_0)$ to exhibit a shared structure across tasks, thereby bypassing the need for accessing all task data during fine-tuning.

\subsection{Function Localization Under Individual Training Constraints}\label{sec:talos}

As the gradient $\nabla_{\bm{\theta}} f(\bm{x},\bm{\theta}_0)$ quantifies the influence of each parameter on the model's output for a given input $\bm{x}$, it serves as a direct measure of \textit{parameter sensitivity}, describing how small variations in each parameter affect the model's input-output behavior. 

Consequently, to satisfy the function localization constraints in Equation \ref{eq:func_loc_constraint}, our goal is to identify those parameters that have minimal impact on the model. In particular, by denoting the $j$-th element of $\bm{\theta} \in \mathbb{R}^m$ as $\bm{\theta}_{[j]}$, we define the \textit{least-sensitive} parameters as the ones for which $\nabla_{\bm{\theta}_{[j]}} f(\bm{x},\bm{\theta}_0) \approx 0$.
We hypothesize that such parameters remain \emph{least sensitive} across all tasks (\ie $\forall \bm{x} \in \mathcal{D}$) and can thus be determined independently of the specific task, without having to access all task data.

To test our hypothesis, we conduct a sensitivity analysis following \citet{chaudhry2018riemannian, pascanu2013revisiting, mergingfisher_2024}.
We define $f(\bm{x},\bm{\theta}_0) \triangleq \log p_{\bm{\theta}_0}(y|\bm{x})$, where $p_{\bm{\theta}_0}(y|\bm{x})$ denotes the probability of assigning class $y$ to $\bm{x}$. To quantify how changes in the parameters influence the model's output, we rely on the Fisher Information matrix (FIM) \citep{fisher1922mathematical, amari1996neural}, a positive semi-definite symmetric matrix given by: 
{\begin{equation*}
    F(\bm{\theta}_0, \mathcal{D}_t) = \mathbb{E}_{\bm{x} \sim \mathcal{D}_t} [ \mathbb{E}_{y \sim p_{\bm{\theta}_0}(y|\bm{x})} [\nabla_{\bm{\theta}} \log p_{\bm{\theta}_0}(y|\bm{x}) \nabla_{\bm{\theta}} \log p_{\bm{\theta}_0}(y|\bm{x})^\top ] ].
\end{equation*}}
For a parameter with index $j \in 1,\dots,m$, the corresponding value on the diagonal of the FIM represents its sensitivity, 
{\begin{equation}\label{eq:diag_fim}
    F_{[j,j]}(\bm{\theta}_0, \mathcal{D}_t) = \frac{1}{N} \sum_{i=1}^N \mathbb{E}_{y \sim p_{\bm{\theta}_0}(y|\bm{x}_i)} [\nabla_{\bm{\theta}_{[j]}} \log p_{\bm{\theta}_0}(y|\bm{x}_i)]^2 ~,
\end{equation}}
where $\bm{x}_1, \ldots, \bm{x}_N \in \mathcal{D}_t$ are \iid examples, while the expectation on the output can be computed via sampling from the distribution of $p_{\bm{\theta}_0}(y|\bm{x}_i)$. The lower $F_{[j,j]}(\bm{\theta}_0, \mathcal{D}_t)$, the less the model will be affected by the $j$-th parameter changes.

\begin{figure}[t!]
    \centering
    \vspace{-5mm}
    \includegraphics[width=\linewidth]{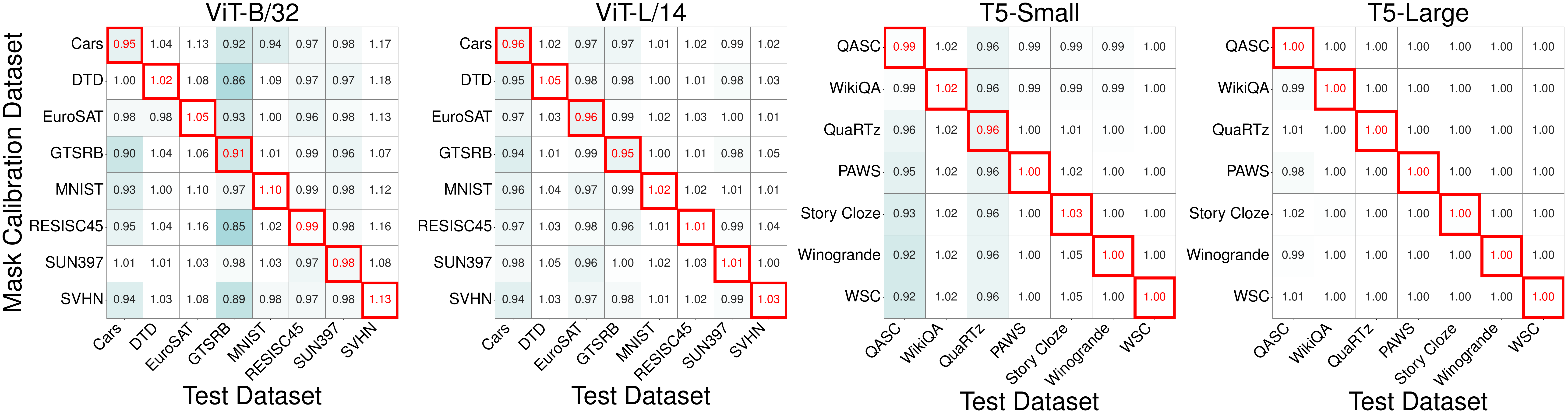}\\
    
    \vspace{-2.5mm}\caption{\textbf{Relative performance when pruning parameters with low sensitivity.} The heatmaps illustrate the effect of pruning the parameters with the lowest sensitivity (measured by $[F_{[j,j]}(\bm{\theta}_0, \mathcal{D}_t)]_{j=1}^m$) on different tasks across various pre-trained models using data from different tasks. Each grid compares the accuracy ratios for models after pruning, where the rows represent the task dataset $\mathcal{D}_t$ used to identify the parameters with the lowest sensitivity, and the columns show the model's zero-shot performance on each task after pruning those parameters. The accuracy ratios are normalized by the model's performance before pruning. The sparsity ratio (10\%) was found as the maximal sparsity that minimally influenced the model's output on the mask calibration dataset.
    }
    \label{fig:shared_bottomk}
    \vspace{-5mm}
\end{figure}

\textbf{Least sensitive parameters are shared across tasks.} To study the role of the least sensitive parameters across tasks, we performed a pruning experiment, illustrated in Figure \ref{fig:shared_bottomk}. We first identified the parameters with the lowest $F_{[j,j]}(\bm{\theta}_0, \mathcal{D}_t)$ by using only data from task $t$.  We then pruned these parameters from the network and evaluated its performance on $t$ and other tasks $t'\neq t$. The results show that the pruned model retains its \textit{zero-shot} performance over all tasks. We conclude that the least sensitive parameters can be effectively identified independently of the specific task, empirically supporting our hypothesis (further validation of this phenomenon and discussion in Appendix \ref{sec:sharing_analysis}). 

Consequently, \emph{function localization} can be achieved by updating only the least sensitive parameters, as for such updates the resulting dot product in Equation \ref{eq:func_loc_constraint} is expected to be minimal across all tasks (we will expand on this in Section \ref{sec:talos_behavior}). Thus, we propose learning task vectors via a selective \longname (\shortname), wherein \emph{only the parameters with the lowest sensitivity are sparsely updated during fine-tuning}.

\subsection{\shortname Implementation}
Sparse fine-tuning consists in introducing a binary mask $\bm{c} \in \{0, 1\}^m$ to control which parameters are updated during gradient descent.
Specifically, at each $i$-th iteration, the update rule becomes:
\begin{equation}
\bm{\theta}^{(i)} = \bm{\theta}^{(i-1)} - \gamma [\bm{c} \odot \nabla_{\bm{\theta}} \mathcal{L}(f(\bm{x},\bm{\theta}^{(i-1)}), y)]~,
\end{equation}
where $\gamma$ is the learning rate, $\mathcal{L}$ is the loss function, and $\odot$ represents the element-wise product.

To achieve function localization we selectively update only the parameters with minimal impact on the model's output. Based on what was discussed earlier, we score each parameter using the diagonal elements of the FIM\footnote{Sensitivity scoring can be implemented through different approaches, as long as they preserve the same ranking as the FIM. For instance, given a scalar output and $f(\bm{x},\bm{\theta}_0) \triangleq \log p_{\bm{\theta}_0}(y|\bm{x})$, $\mathbb{E}_{\bm{x}}[|\nabla_{\bm{\theta}}f(\bm{x},\bm{\theta}_0)|]$ yields the same ranking as the diagonal of the FIM $\mathbb{E}_{\bm{x}}[[\nabla_{\bm{\theta}} \log p_{\bm{\theta}_0}(y|\bm{x}_i)]^2]$, as the absolute value function ($h(x)=|x|$) and the squaring function ($h(x)=x^2$) are both monotonically increasing in the interval $]0,+\infty[$.} $\bm{s} = [F_{[j,j]}(\bm{\theta}_0, \mathcal{D}_t)]_{j=1}^m \in \mathbb{R}^m$ and sort them to identify the index $j^*$ of the $k$-th lowest element in $\bm{s}$. This value is adopted as a threshold and we set  $\bm{c}_{[j]} = 0$ if $\bm{s}_{[j]} \geq \bm{s}_{[j^*]}$ effectively freezing these parameters. Otherwise, $\bm{c}_{[j]} = 1$, allowing these parameters to be updated during fine-tuning. 
Note that the estimation of $\bm{c}$ may be susceptible to gradient noise \citep{tanaka2020synflow}. Thus, we follow standard Pruning-at-Initialization practices \citep{tanaka2020synflow} and iteratively refine $\bm{c}$ in multiple rounds (we provide full details of \shortname, alongside its pseudocode in Appendix \ref{sec:mask_calibration}).

\subsection {Insights on Sparsity and Linear Behavior}\label{sec:talos_behavior}
\textbf{\shortname promotes linear behavior.} Parameters with the smallest (ideally near-zero) $F_{[j,j]}(\bm{\theta}_0, \mathcal{D}_t)$ are associated with flatter regions in the loss landscape, as for $\bm{\theta}_0$ the FIM equals the Gauss-Newton approximation of the Hessian \citep{pennington2018spectrum, kunstner2019limitations}. Updating parameters in a flat subspace allows the gradient to be approximately constant throughout fine-tuning, a necessary condition for operating in the linearized regime \citep{malladi2023kernel}. This means that fine-tuning the least sensitive parameters \emph{inherently promotes a linear behavior} without requiring explicitly linearizing the network. We follow \citet{Tangent_task_arith_2023} to confirm this claim in Appendix \ref{sec:fixed_features}.

\textbf{Function localization in \shortname.} Given the \emph{least sensitive} parameters are shared across tasks, the function localization constraints of Equation \ref{eq:func_loc_constraint} for \shortname can be rewritten and upper bounded as
\begin{equation}
\label{eq:sparsity_bound}
\forall \bm{x} \in \mathcal{D}_t,~ |\bm{c} \odot (\bm{\tau}_t^\top \nabla_{\bm{\theta}} f(\bm{x},\bm{\theta}_0)| \leq \| \bm{c} \odot \bm{\tau}_t\| \cdot \max_{\bm{x} \in \mathcal{D}_t} \| \bm{c} \odot \nabla_{\bm{\theta}} f(\bm{x},\bm{\theta}_0)\| \leq k^2 \cdot \mu \cdot \eta~.
\end{equation}
Here, $\eta = \max_{\bm{x}} |\nabla_{\bm{\theta}_{[j^*]}} f(\bm{x},\bm{\theta}_0)|$ is the magnitude of the $k$-th largest gradient element, capturing the maximum sensitivity of the fine-tuned parameters to input data. $\mu = \max_{j} |\bm{c}_{[j]} \odot \bm{\tau}_{t_{[j]}}|$ represents the maximum change in any of the updated parameters during fine-tuning. 
Inequality \ref{eq:sparsity_bound} provides an upper bound on the degree of \emph{function localization} of  $\bm{\tau}_t$ obtained via \shortname. Having this quantity equal zero ensures no task interference, as the overall output falls back to $f(\cdot, \bm{\theta}_0)$ which by definition is \emph{weight disentangled}. Yet, this means that no learning has occurred. Instances of this are when no parameter is updated ($k=0$) or when only parameters with exactly zero influence ($\eta = 0$) are fine-tuned. Apart from these cases, fine-tuning the least sensitive parameters allows for a minimal increase of this bound while still allowing to learn the task, as even parameters with marginal influence can collectively contribute to task performance \citep{ben-zaken-etal-2022-bitfit, xu2020one, liao-etal-2023-parameter} (in Appendix \ref{sec:single_task_perf_peft} we show that \shortname enables learning on par with other PEFT baselines). Indeed, as the model is robust to changes within the flat subspace defined by its least sensitive parameters, learning $\bm{\tau}_t$ in this subspace ensures minimal impact on the model's output for other tasks as well (we empirically validate this in Figure \ref{fig:heatmaps_alignment}).
As detailed in Appendix \ref{sec:implementation}, $k$ is a hyperparameter controlling the sparsity ratio of $\bm{c}$, thus, indirectly controlling the degree of function localization. We tuned it at the task level, resulting in optimal sparsity ratios between 90\% and 99\% (ablation in Appendix \ref{sec:k_ablation}).

\section{Experiments}

Our experimental evaluation focuses on the established Task Arithmetic framework outlined by \citet{ilharco2022patching, ilharco2023editing}, specifically targeting Task Addition and Task Negation, encompassing both language and vision domains. 
In the following we describe the baselines we compared our \shortname against. Further details regarding the experimental setups, the relevant metrics, the implementation of the experiments, as well as the data and architectures used, are deferred to Appendix \ref{sec:implementation}. Additionally, in Appendix \ref{sec:talos+modelmerging} we test different model merging schemes on task vectors obtained with \shortname.

\textbf{Baselines.}
We consider three families of methods as references. (i) \textbf{Full fine-tuning} methods aim to produce task vectors $\bm{\tau}_t$ by fine-tuning all the parameters of the network. Specifically, \emph{Non-linear fine-tuning} (FT) \citep{ilharco2022patching, ilharco2023editing} minimizes a standard cross-entropy loss, while \emph{Linearized FT} fine-tunes the linearized counterpart of the network, as in \cite{Tangent_task_arith_2023}.
(ii) \textbf{Post-hoc} methods refine $\bm{\tau}_t$ after it has been obtained via fine-tuning (as prescribed by the respective methods, we apply these post-hoc approaches on non-linear FT checkpoints). \emph{TIES-Merging} \citep{yadav2023ties} reduces redundancy in $\bm{\tau}_t$ by magnitude pruning, keeping only the top-$k$ highest magnitude parameters, and addressing sign conflicts when merging task vectors. \emph{TALL Mask / Consensus} \citep{wang2024consensus} identifies task-specific parameters in $\bm{\tau}_t$ by comparing them to the sum of task vectors. It then merges multiple task vectors by using an element-wise OR operation between masks to further identify and remove conflicting parameters. \emph{DARE} \citep{dareSupermario} randomly sparsifies $\bm{\tau}_t$ to eliminate redundancy and upweights the remaining parameters based on the percentage that was removed. \emph{Breadcrumbs} \citep{davari2023model} reduces redundancy using magnitude pruning and eliminates weight outliers within the retained top-$k$ parameters.
Although these methods have been presented for task addition, we also test their ability of handling task negation.
(iii) \textbf{Parameter-efficient fine-tuning (PEFT)} methods aim to obtain task vectors by efficiently fine-tuning the network, using far fewer resources compared to full fine-tuning. We compare against \emph{L-LoRA} \citep{tang2023parameter}, which applies linearized low-rank adapters to the $\bm{Q}$ and $\bm{V}$ projections in self-attention layers. This approach was specifically designed for Task Arithmetic and offers superior performance over standard LoRA.
For sparse fine-tuning, we use \emph{LoTA} \citep{panda2024lottery}, a method that leverages the Lottery Ticket hypothesis \citep{frankle2018lottery} to select the top-$k$ parameters when sparsely fine-tuning the network, making it suitable for model merging.

\subsection{Task Arithmetic Results}

We thoroughly evaluate \shortname on its ability to derive task vectors that enable model editing through simple arithmetic operations on model parameters.

\textbf{Task Addition.} In this benchmark, the sum of the task vectors $\sum_{t} \alpha_t \bm{\tau}_t$ is added to a pre-trained checkpoint to produce a multi-task model $f(\cdot, \bm{\theta}_0 + \sum_{t} \alpha_t \bm{\tau}_t)$. The success is measured in terms of the maximum average accuracy over the different tasks. As done by \citet{Tangent_task_arith_2023, tang2023parameter}, we also report the average normalized accuracy over the tasks. The normalization is performed with respect to the single-task accuracies achieved by the model fine-tuned on each task (see Appendix \ref{sec:implementation}).
The results in Table \ref{tab:task_addition} demonstrate the effectiveness of our proposed method across various model scales and modalities. \shortname consistently outperforms existing approaches, with evident improvements in normalized accuracy of 1.88\% to 4.65\% over the second best method across all model variants.
Such a metric provides insights into the outstanding ability of \shortname to maximize the benefits of model combination while mitigating interference.

For vision models, \shortname exhibits strong performance across all scales, with absolute accuracy gains of up to 2.61\% over the closest competitor. In NLP, \shortname maintains its leading position, although the gains are less striking than in vision experiments. Nevertheless, the improvements are particularly pronounced in larger models, suggesting that \shortname scales well with model size. Notably, \shortname's performance surpasses both full fine-tuning and post-hoc methods across the board. This suggests that our parameter-efficient approach can achieve superior results while potentially reducing computational costs, a crucial factor when working with large-scale models.

\textbf{Task Negation.} In this benchmark a task vector $\bm{\tau}_t$ is subtracted from the pre-trained checkpoint to reduce the performance on task $t$, producing the model $f(\cdot, \bm{\theta}_0 - \alpha_t \bm{\tau}_t)$. By following \citet{Tangent_task_arith_2023}, the success is measured in terms of the maximum drop in accuracy on the forgetting task that retains at least 95\% of the accuracy on the control task. Results are averaged over tasks and presented in Table \ref{tab:task_negation}.
For vision models, \shortname achieves the lowest target task accuracies while maintaining high control task performance, indicating superior ability to selectively remove targeted task information. For T5 models, all methods, including \shortname, face significant challenges in Task Negation. The results show a much tighter clustering of performance across different approaches. This suggests that negating specific language tasks without substantially impacting the control task accuracy is inherently more difficult than in vision models. Despite this challenge, \shortname still manages to achieve the best balance between target and control task performance.

\begin{table}[t!]
    \vspace{-5mm}
    \centering
    \setlength{\aboverulesep}{0pt}
    \setlength{\belowrulesep}{0pt}
    \setlength{\extrarowheight}{.75ex}
    \resizebox{.98\linewidth}{!}{\begin{tabular}{l a a b b a a | b b a a b b}%
    \toprule

    \multicolumn{1}{c}{Method} & \multicolumn{2}{a}{ViT-B/32} & \multicolumn{2}{b}{ViT-B/16} & \multicolumn{2}{a|}{ViT-L/14} & \multicolumn{2}{b}{T5-Small} & \multicolumn{2}{a}{T5-Base} & \multicolumn{2}{b}{T5-Large} \\
    {} & Abs. ($\uparrow$) & Norm. ($\uparrow$) & Abs. ($\uparrow$) & Norm. ($\uparrow$) & Abs. ($\uparrow$) & Norm. ($\uparrow$) & Abs. ($\uparrow$) & Norm. ($\uparrow$) & Abs. ($\uparrow$) & Norm. ($\uparrow$) & Abs. ($\uparrow$) & Norm. ($\uparrow$) \\
    \midrule

    Pre-trained (Zero-shot) & 47.72 & - & 55.83 & - & 65.47 & - & 55.70 & - & 53.51 & - & 51.71 & - \\
    \midrule

    \multicolumn{1}{c}{\tb{Full Fine-tuning Methods}} \\
    \midrule
    
    Non-linear FT \citep{ilharco2023editing} & 71.25 & 76.94 & 72.85 & 77.17 & 86.09 & 90.14 & \tb{65.04} & 87.98 & 74.20 & 90.63 & 75.37 & 85.25 \\
    Linearized FT \citep{Tangent_task_arith_2023} & 76.70 & 85.86 & 80.01 & \ul{87.29} & \ul{88.29} & \ul{93.01} & 64.13 & 86.62 & \ul{74.69} & 92.12 & 69.38 & 78.95 \\
    \midrule

    \multicolumn{1}{c}{\tb{Post-hoc Methods}} \\
    \midrule

    TIES-Merging \citep{yadav2023ties} & 74.79 & 82.84 & 77.09 & 82.13 & 88.16 & 92.56 & 62.53 & 94.83 & 70.74 & 92.37 & 74.30 & 86.36 \\
    TALL Mask / Consensus \citep{wang2024consensus} & 74.55 & 80.27 & 74.92 & 79.12 & 86.89 & 90.81 & 63.61 & \ul{95.34} & 73.31 & 91.60 & \ul{77.31} & 87.84 \\
    DARE \citep{dareSupermario} & 70.88 & 76.59 & 73.08 & 77.51 & 85.95 & 90.04 & 63.89 & 89.09 & 74.26 & 91.49 & 76.20 & 86.51 \\
    Breadcrumbs \citep{davari2023model} & 69.39 & 79.51 & 71.93 & 78.94 & 84.78 & 92.97 & 61.19 & 92.23 & 73.89 & \ul{92.70} & 73.41 & 87.07 \\
    \midrule

    \multicolumn{1}{c}{\tb{Parameter-efficient Fine-tuning Methods}} \\
    \midrule
    
    L-LoRA \citep{tang2023parameter} & \ul{78.00} & \ul{86.08} & \ul{80.61} & 85.83 & 87.77 & 91.87 & 60.29 & 94.46 & 68.76 & 91.98 & 72.10 & 87.78 \\
    LoTA \citep{panda2024lottery} & 64.94 & 74.37 & 79.11 & 83.97 & 87.66 & 91.69 & \ul{64.21} & 87.92 & 74.31 & 92.25 & 75.84 & \ul{88.14} \\

    \textbf{\shortname (Ours)} & \tb{79.67} \up{1.67} & \tb{90.73} \up{4.65} & \tb{82.60} \up{1.99} & \tb{91.41} \up{4.12} & \tb{88.37} \up{0.08} & \tb{95.20} \up{2.19} & \tb{65.04} \na{0.00} & \tb{97.22} \up{1.88} & \tb{75.93} \up{1.24} & \tb{95.87} \up{3.17} & \tb{79.07} \up{1.76} & \tb{90.61} \up{2.47} \\
    
    \bottomrule
    \end{tabular}%
    }
    \caption{\textbf{Task Addition results.} Average absolute accuracies (\%) and normalized accuracies (\%) of different CLIP ViTs and T5 pre-trained models edited by adding task vectors on each of the downstream tasks. We normalize performance of each method by their single-task accuracy. 
    \textbf{Bold} indicates the best results. \underline{Underline} the second best.}
    \label{tab:task_addition}
    \vspace{-3mm}
\end{table}

\begin{table}[t!]
    \centering
    \setlength{\aboverulesep}{0pt}
    \setlength{\belowrulesep}{0pt}
    \setlength{\extrarowheight}{.75ex}
    \resizebox{.98\linewidth}{!}{\begin{tabular}{l a a b b a a | b b a a b b}%
    \toprule

    \multicolumn{1}{c}{Method} & \multicolumn{2}{a}{ViT-B/32} & \multicolumn{2}{b}{ViT-B/16} & \multicolumn{2}{a|}{ViT-L/14} & \multicolumn{2}{b}{T5-Small} & \multicolumn{2}{a}{T5-Base} & \multicolumn{2}{b}{T5-Large} \\
    {} & Targ. ($\downarrow$) & Cont. ($\uparrow$) & Targ. ($\downarrow$) & Cont. ($\uparrow$) & Targ. ($\downarrow$) & Cont. ($\uparrow$) & Targ. ($\downarrow$) & Cont. ($\uparrow$) & Targ. ($\downarrow$) & Cont. ($\uparrow$) & Targ. ($\downarrow$) & Cont. ($\uparrow$) \\
    \midrule

    Pre-trained (Zero-shot) & 47.72 & 63.26 & 55.83 & 68.37 & 65.47 & 75.53 & 55.70 & 45.70 & 53.51 & 45.30 & 51.71 & 45.70 \\
    \midrule

    \multicolumn{1}{c}{\tb{Full Fine-tuning Methods}} \\
    \midrule
    
    Non-linear FT \citep{ilharco2023editing} & 24.04 & 60.36 & 20.36 & 64.79 & 20.61 & 72.72 & 43.06 & \ul{45.47} & \ul{40.06} & 45.16 & 41.54 & 45.49 \\
    Linearized FT \citep{Tangent_task_arith_2023} & \ul{11.20} & 60.74 & \ul{10.97} & 65.55 & \ul{10.86} & 72.43 & 44.47 & 44.94 & 40.16 & \ul{45.27} & 41.37 & \tb{45.70} \\
    \midrule

    \multicolumn{1}{c}{\tb{Post-hoc Methods}} \\
    \midrule

    TIES-Merging \citep{yadav2023ties} & 21.94 & \tb{61.49} & 19.72 & \ul{65.69} & 24.50 & \ul{73.41} & 55.01 & 45.30 & 40.30 & 45.13 & 46.19 & 45.56 \\
    TALL Mask / Consensus \citep{wang2024consensus} & 23.31 & 60.54 & 20.71 & 65.17 & 22.33 & 73.30 & 43.43 & 45.41 & 40.14 & 45.20 & \ul{41.26} & 45.59 \\
    DARE \citep{dareSupermario} & 25.04 & 60.60 & 22.22 & 64.98 & 20.94 & 72.66 & \ul{42.53} & 45.36 & 40.24 & 45.16 & 41.29 & \tb{45.70} \\
    Breadcrumbs \citep{davari2023model} & 24.27 & 60.58 & 21.60 & 65.22 & 20.69 & 72.95 & 53.03 & 45.19 & 40.46 & 45.14 & 41.49 & 45.51 \\
    \midrule

    \multicolumn{1}{c}{\tb{Parameter-efficient Fine-tuning Methods}} \\
    \midrule
    
    L-LoRA \citep{tang2023parameter} & 17.29 & 60.75 & 19.33 & \ul{65.69} & 19.39 & 73.14 & 55.30 & 45.24 & 51.33 & 45.10 & 48.37 & 45.51 \\
    LoTA \citep{panda2024lottery} & 21.09 & \ul{61.01} & 17.76 & 65.60 & 22.11 & 73.21 & 54.70 & 45.13 & 40.50 & 45.24 & 44.33 & 45.47 \\
    %\midrule

    \textbf{\shortname (Ours)} & \tb{11.03} \up{0.17} & 60.69 \dw{0.80} & \tb{10.58} \up{0.39} & \tb{66.11} \up{0.42} & \tb{10.68} \up{0.18} & \tb{73.63} \up{0.22} & \tb{39.64} \up{2.89} & \tb{45.67} \up{0.20} & \tb{38.49} \up{1.57} & \tb{45.28} \up{0.01} & \tb{37.20} \up{4.06} & \tb{45.70} \na{0.00} \\
    
    \bottomrule
    \end{tabular}%
    }
    %\vspace{1.5mm}
    \caption{\textbf{Task Negation results.} Average minimal accuracy (\%) of different CLIP ViTs and T5 pre-trained models edited by subtracting a task vector from a target task while retaining at least $95\%$ of their performance on the control task. We average the minimal accuracy over each of the downstream tasks. %For each of the families of methods, 
    \textbf{Bold} indicates the best results. \underline{Underline} the second best.}
    \label{tab:task_negation}
    \vspace{-3mm}
\end{table}

\subsection{Weight Disentanglement and Localization}\label{sec:validation}
\begin{figure}[t!]
    \vspace{-5mm}
    \centering
    \includegraphics[width=0.87\linewidth]{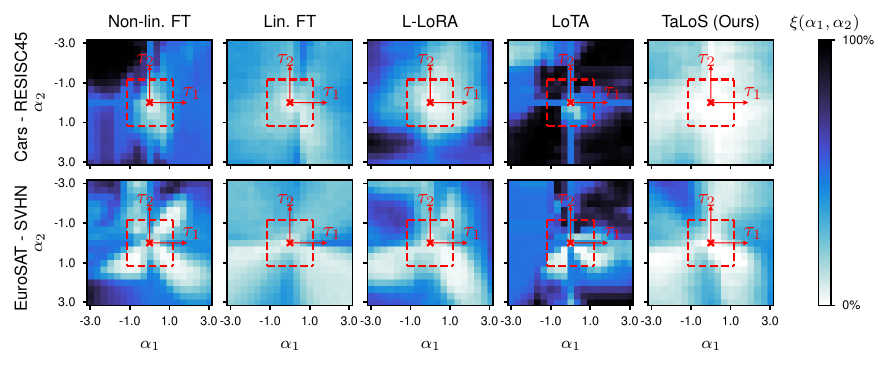}
    \includegraphics[width=0.87\linewidth]{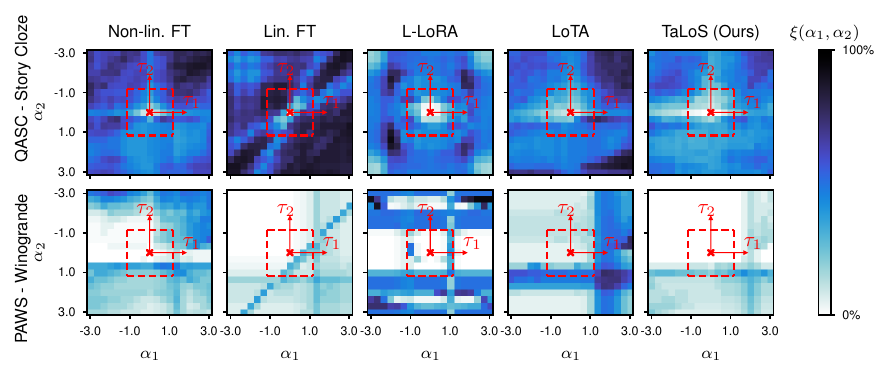}
    
    \vspace{-3.5mm}\caption{\textbf{Visualizing weight disentanglement error.} The heatmaps illustrate the disentanglement error $\xi(\alpha_1, \alpha_2)$ of each fine-tuning strategy on both a CLIP ViT-B/32 model (top) and a T5-Small model (bottom) across two task pairs. Lighter areas highlight regions of the weight space where disentanglement is more pronounced. The red box indicates the search space within which the optimal $\alpha$ values were searched (refer to Appendix \ref{sec:implementation}). 
    We chose the task pairs to visualize by following \citet{Tangent_task_arith_2023} for vision and a criterion akin to the one used in \citet{tang2023parameter} for language. 
    %On vision, we choose the task pairs to visualize based on the choices of \citet{Tangent_task_arith_2023}. On language, we follow a criterion akin to the one used in \citet{tang2023parameter}.
    }
    \label{fig:weight_disentangle}
    \vspace{-3mm}
\end{figure}

The improved localization provided by \shortname seems to play a crucial role in driving effective task arithmetic. Here we delve deeper into this aspect with tailored analyses. First, we assess how well the weight disentanglement property holds. Then, for each training recipe, we evaluate the degree of task component localization on each task.

\textbf{Weight disentanglement error visualization.} \citet{Tangent_task_arith_2023, tang2023parameter} proposed to evaluate the \emph{disentanglement error} defined as 
\begin{equation}\label{eq:disentanglement_err}
   \xi(\alpha_1, \alpha_2) = \sum_{t=1}^2 \mathbb{E}_{\bm{x} \in \mathcal{D}_t}[\text{dist}(f(\bm{x}, \bm{\theta}_0 + \alpha_1\bm{\tau}_1), f(\bm{x}, \bm{\theta}_0 + \alpha_1\bm{\tau}_1 + \alpha_2\bm{\tau}_2))] 
\end{equation}
where the \emph{prediction error} $\text{dist}(y_1, y_2) = \mathds{1}(y_1 \neq y_2)$ is taken as the distance metric. Generally, given a pair $(\alpha_1, \alpha_2)$, the smaller the value of $\xi(\alpha_1, \alpha_2)$ the more weight disentangled a model is. Maintaining a low disentanglement error as $\alpha_1$ and $\alpha_2$ increase provides an even stronger evidence of the weight disentanglement property.

In Figure \ref{fig:weight_disentangle}, we report $\xi(\alpha_1, \alpha_2)$ across different fine-tuning strategies for both the CLIP ViT-B/32 and T5-Small models on two task pairs.
Overall there is a clear difference in disentanglement patterns between vision and language models. For the latter, the patterns are more consistent across strategies, which may explain why the differences in task arithmetic performance are notable in vision experiments and less pronunced in language experiments (ref. to Tables \ref{tab:task_addition}, \ref{tab:task_negation}).

By focusing on vision models we observe that Linearized FT, L-LoRA, and our approach demonstrate improved disentanglement (indicated by lighter regions) than non-linear fine-tuning, with our method performing the best overall.  We remind that L-LoRA approximate the behavior of Linearized FT via adapters but still lacks to optimize the task localization property. 
Interestingly, LoTA shows a much lower degree of disentanglement. We remark that this approach selects and updates task-specific parameters while \shortname focuses on task-generic ones and this difference accounts for the observed behavior. 

For language, Linearized FT and L-LoRA yield mixed results depending on the pairs of considered tasks. LoTA seems able to improve over non-linearized FT but with different extents across tasks and it is consistently outperformed by \shortname. 

\begin{figure}[t!]
    \centering
    \vspace{-5mm}
    \includegraphics[width=\linewidth]{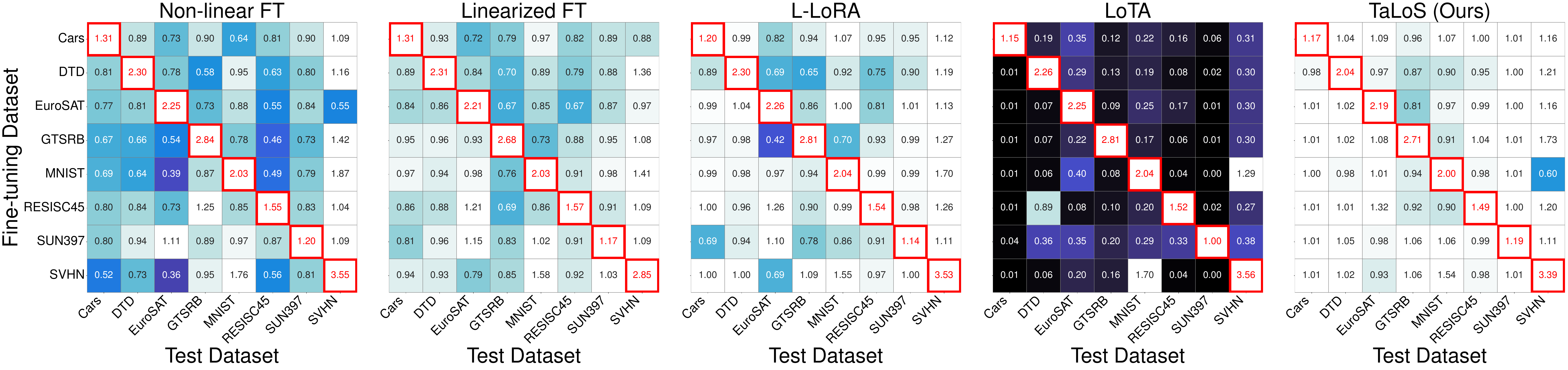}\\
    \vspace{12pt}
    \includegraphics[width=\linewidth]{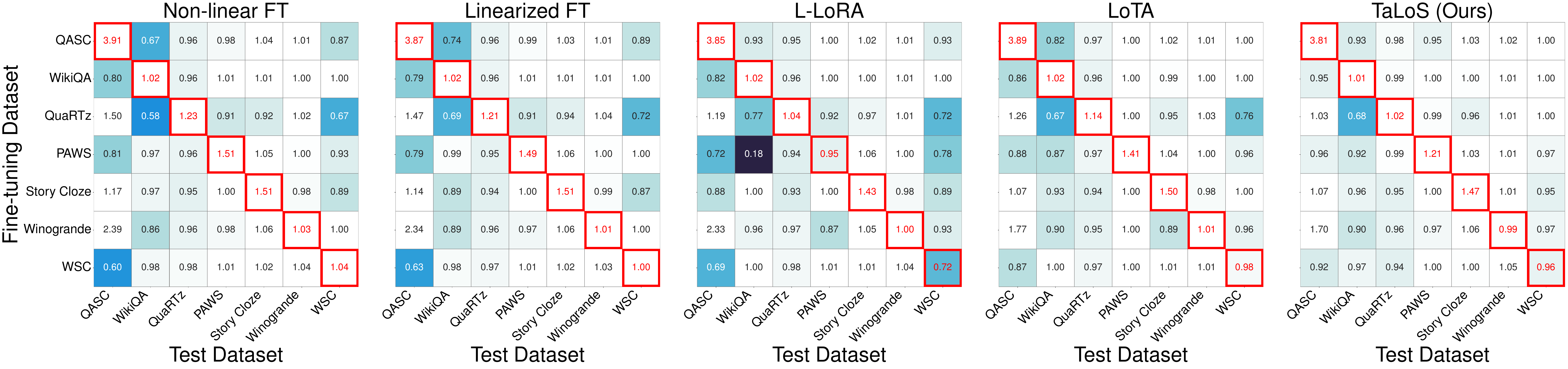}
    
    \vspace{-2.5mm}\caption{\textbf{Function localization.} The heatmaps present the accuracy ratios for fine-tuned models across tasks for CLIP ViT-B/32 (top) and T5-Small (bottom) models. Each row indicates a model fine-tuned on a specific task, with columns representing its performance on different test datasets. Accuracy ratios are normalized by the pre-trained model's performance. Lighter colors indicate better performance, suggesting minimal interference between the fine-tuned model and other tasks' input spaces. The red diagonal highlights each model's test performance on its specific fine-tuning task.}
    \label{fig:heatmaps_alignment}
    \vspace{-3mm}
\end{figure}

\textbf{Function localization.} We experimentally assess the function localization property of \shortname by comparing it with other fine-tuning methods. From the definition in Equation \ref{eq:func_loc_constraint}, we know that when this property holds, each task activates only for its specific data support. Thus, we should observe an advantage in the prediction output when testing on that task, and the same performance of the pre-trained model for all the others tasks. 
Figure \ref{fig:heatmaps_alignment} confirms the expected behavior for \shortname in vision, while the competitors display more interference between tasks, as indicated by darker hues off the diagonal.
Interestingly, for NLP tasks all methods exhibit natural function localization, as reflected by the lighter regions in the figure. This provides us the opportunity to remark the importance of extensive model analysis as conclusions drawn from a single domain where linearization is sufficient might be misleading. 

\subsection{Weight Sparsity Structure and Efficiency}

\begin{figure}[t!]
    \centering
    \includegraphics[width=\linewidth]{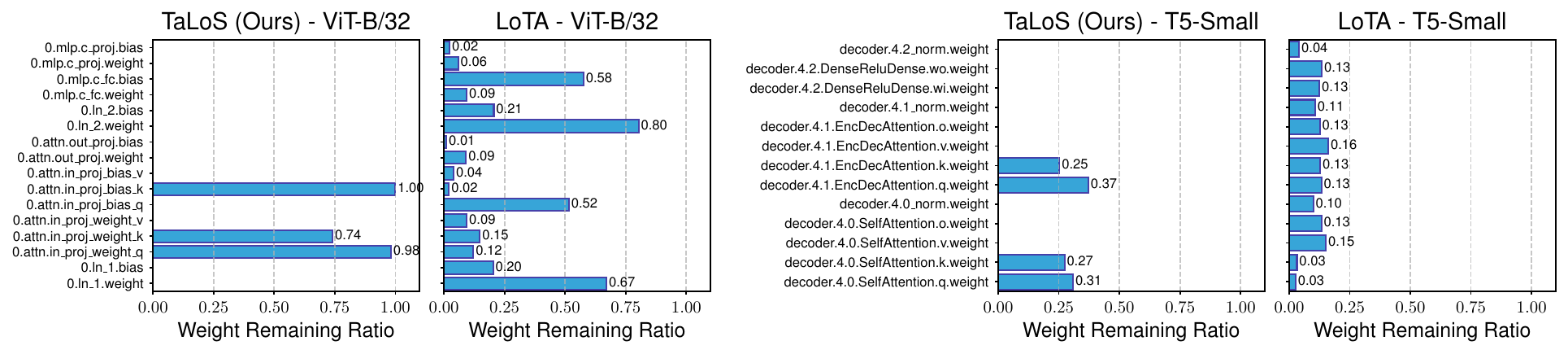}
    
    \vspace{-2.5mm}\caption{\textbf{Visualization of mask calibration.} Percentage of parameters selected for sparse fine-tuning in a transformer block of a ViT-B/32 (left) and a T5-Small (right) models, after our method's mask calibration vs. LoTA's mask calibration, at 90\% sparsity. On ViT-B/32, we calibrate the masks on the Cars dataset \citep{krause2013cars}, while on T5-Small we use QASC \citep{khot2020qasc}. Full visualizations of all masked layers are reported in Appendix \ref{sec:full_mask_vis}.}
    \label{fig:sparsity_block}
    \vspace{-3mm}
\end{figure}

\begin{table}[t!]
    \vspace{-5mm}
    \centering
    \setlength{\aboverulesep}{0pt}
    \setlength{\belowrulesep}{0pt}
    \setlength{\extrarowheight}{.75ex}
    \resizebox{.98\linewidth}{!}{\begin{tabular}{l | a b | a | b || a a b b}%
    \toprule

    \multicolumn{1}{c|}{Method} & \multicolumn{4}{b||}{Effective Cost of Fine-tuning} & \multicolumn{2}{a}{Task Addition} & \multicolumn{2}{b}{Task Negation} \\
    \cline{2-5}
    
    {} & Forward-Backward Pass Time (s) & Optim. Step Time (s) & Tot. Iteration Time (s) & Peak Memory Usage (GiB) & Abs. ($\uparrow$) & Norm. ($\uparrow$) & Targ. ($\downarrow$) & Cont. ($\uparrow$) \\
    \midrule
    
    \multicolumn{1}{c}{\tb{ViT-B/32}} \\
    \midrule
    Non-linear FT \citep{ilharco2023editing} & 0.3608 \footnotesize{$\pm$ 0.0036} & \ul{0.0114} \footnotesize{$\pm$ 0.0010} & 0.3722 \footnotesize{$\pm$ 0.0037} & 6.5 & 71.25 & 76.94 & 24.04 & 60.36 \\
    Linearized FT \citep{Tangent_task_arith_2023} & 0.6858 \footnotesize{$\pm$ 0.0042} & 0.0103 \footnotesize{$\pm$ 0.0020} & 0.6961 \footnotesize{$\pm$ 0.0047} & 10.2 & 76.70 & 85.86 & \ul{11.20} & 60.74 \\
    \hline\hline
    L-LoRA \citep{tang2023parameter} & \ul{0.3270} \footnotesize{$\pm$ 0.0076} & \tb{0.0036} \footnotesize{$\pm$ 0.0032} & \ul{0.3306} \footnotesize{$\pm$ 0.0082} & \ul{5.3} & \ul{78.00} & \ul{86.08} & 17.29 & \ul{60.75} \\
    LoTA \citep{panda2024lottery} & 0.3289 \footnotesize{$\pm$ 0.0041} & 0.1269 \footnotesize{$\pm$ 0.0050} & 0.4558 \footnotesize{$\pm$ 0.0065} & 6.8 & 64.94 & 74.37 & 21.09 & \tb{61.01} \\
    %\midrule
    \textbf{\shortname (Ours)} & \tb{0.1256} \footnotesize{$\pm$ 0.0045} & 0.0388 \footnotesize{$\pm$ 0.0040} & \tb{0.1644} \footnotesize{$\pm$ 0.0060} & \tb{4.7} & \tb{79.67} & \tb{90.73} & \tb{11.03} & 60.69 \\
    \midrule
    
    \multicolumn{1}{c}{\tb{ViT-L/14}} \\
    \midrule
    Non-linear FT \citep{ilharco2023editing} & 1.2174 \footnotesize{$\pm$ 0.0097} & \ul{0.0156} \footnotesize{$\pm$ 0.0055} & 1.2330 \footnotesize{$\pm$ 0.0112} & 18.6 & 86.09 & 90.14 & 20.61 & 72.72 \\
    Linearized FT \citep{Tangent_task_arith_2023} & 1.6200 \footnotesize{$\pm$ 0.0067} & 0.0262 \footnotesize{$\pm$ 0.0082} & 1.6462 \footnotesize{$\pm$ 0.0106} & 21.3 & \ul{88.29} & \ul{93.01} & \ul{10.86} & 72.43 \\
    \hline\hline
    L-LoRA \citep{tang2023parameter} & 0.5153 \footnotesize{$\pm$ 0.0077} & \tb{0.0082} \footnotesize{$\pm$ 0.0015} & \ul{0.5235} \footnotesize{$\pm$ 0.0078} & \ul{9.7} & 87.77 & 91.87 & 19.39 & 73.14 \\
    LoTA \citep{panda2024lottery} & 0.8438 \footnotesize{$\pm$ 0.0052} & 0.4449 \footnotesize{$\pm$ 0.0074} & 1.2887 \footnotesize{$\pm$ 0.0090} & 15.4 & 87.66 & 91.69 & 22.11 & \ul{73.21} \\
    %\midrule
    \textbf{\shortname (Ours)} & \tb{0.1891} \footnotesize{$\pm$ 0.0039} & 0.1372 \footnotesize{$\pm$ 0.0036} & \tb{0.3263} \footnotesize{$\pm$ 0.0053} & \tb{7.8} & \tb{88.37} & \tb{95.20} & \tb{10.68} & \tb{73.63} \\
    \midrule

    \multicolumn{1}{c}{\tb{T5-Large}} \\
    \midrule
    Non-linear FT \citep{ilharco2023editing} & 0.9047 \footnotesize{$\pm$ 0.0068} & 0.0894 \footnotesize{$\pm$ 0.0034} & 0.9941 \footnotesize{$\pm$ 0.0076} & 30.0 & 75.37 & 85.25 & 41.54 & 45.49 \\
    Linearized FT \citep{Tangent_task_arith_2023} & 1.7683 \footnotesize{$\pm$ 0.0084} & 0.1170 \footnotesize{$\pm$ 0.0060} & 1.8853 \footnotesize{$\pm$ 0.0103} & 35.1 & 69.38 & 78.95 & \ul{41.37} & \tb{45.70} \\
    \hline\hline
    L-LoRA \citep{tang2023parameter} & \ul{0.7452} \footnotesize{$\pm$ 0.0084} & \tb{0.0136} \footnotesize{$\pm$ 0.0029} & \ul{0.7588} \footnotesize{$\pm$ 0.0089} & \ul{18.2} & 72.10 & 87.78 & 48.37 & 45.51 \\
    LoTA \citep{panda2024lottery} & 0.8526 \footnotesize{$\pm$ 0.0043} & 0.3842 \footnotesize{$\pm$ 0.0019} & 1.2368 \footnotesize{$\pm$ 0.0047} & 32.1 & \ul{75.84} & \ul{88.14} & 44.33 & 45.47 \\
    %\midrule
    \textbf{\shortname (Ours)} & \tb{0.4358} \footnotesize{$\pm$ 0.0075} & \ul{0.0509} \footnotesize{$\pm$ 0.0046} & \tb{0.4867} \footnotesize{$\pm$ 0.0088} & \tb{12.1} & \tb{79.07} & \tb{90.61} & \tb{37.20} & \tb{45.70} \\
    
    \bottomrule
    \end{tabular}%
    }
    %\vspace{1.5mm}
    \caption{\textbf{Computational cost and memory footprint of fine-tuning.} Average iteration time (in seconds) and peak memory usage (in Gibibytes) of different fine-tuning approaches on CLIP ViT-B/32, ViT-L/14 and T5-Large models, alongside their performance on the task arithmetic benchmark. To improve granularity, we report also the average forward-backward time of a single iteration and the average step time of the optimizer. We separate full fine-tuning methods from parameter-efficient fine-tuning methods. Further details on the resource monitoring process can be found in Appendix \ref{sec:implementation}.
    \textbf{Bold} indicates the best results. \underline{Underline} the second best.}
    \label{tab:efficiency}
    \vspace{-3mm}
\end{table}

\textbf{Visualizing task vector masks.} To understand the nature of our sparse fine-tuning approach, we analyze the structure of the masks $\bm{c}$ calibrated using \shortname and compare it with the ones produced by LoTA. Figure \ref{fig:sparsity_block} provides a visualization of the layer-wise percentage of parameters selected for sparse fine-tuning in a transformer block of a ViT-B/32 and a T5-Small models.
The results reveal distinct patterns in parameter selection between \shortname and LoTA across both models. \shortname exhibits a highly structured selection, predominantly preserving parameters in the multihead self-attention layer, particularly in the $\bm{Q}$ and $\bm{K}$ projections. In contrast, LoTA's selection appears more distributed across different layers of the transformer block. Interestingly, our analysis reveals some notable contrasts with L-LoRA \citep{tang2023parameter}, a method specifically designed for task arithmetic. While L-LoRA arbitrarily fine-tunes the $\bm{Q}$ and $\bm{V}$ projections, our findings suggest that, generally, $\bm{Q}$ and $\bm{K}$ play a more significant role in task arithmetic than $\bm{V}$ in the multihead self-attention layers. Additionally, for CLIP ViT-B/32 biases also seemingly play a crucial role for function localization.
This structured sparsity not only provides insights into our method's mask calibration mechanism but also hints at potential efficiency gains, which we explore further in the following.

\textbf{Computational cost and memory footprint.} 
The observed structured sparsity pattern of \shortname suggests that it also provides a highly efficient task arithmetic fine-tuning strategy. To verify it we performed a comparative analysis of the computational cost and memory footprint of \shortname against several fine-tuning methods. 

In Table \ref{tab:efficiency} we present the collected time and memory costs with detailed average time (in seconds) for a single training iteration's forward and backward pass. This is separated because approaches like Linearized FT and L-LoRA involve specialized forward passes that require Jacobian-vector products with respect to LoTA and \shortname, which operate similarly to non-linear FT. We also report the time (in seconds) spent by the optimizer updating parameters, as LoTA and \shortname require an additional mask-based element-wise multiplication to prevent updates to certain parameters by masking gradients.
Additionally, we provide the total time (sum of these two values) and the peak memory usage (in Gibibytes) recorded during fine-tuning. Overall, the ability to freeze a large number of parameters, thanks to well-structured mask sparsity of our approach improves the total iteration time. Although our method has a slower optimizer step compared to other approaches, the faster forward-backward pass compensates, making \shortname the leading method. In terms of memory usage, the benefits are especially notable for large models, where only a small subset of parameters requires fine-tuning, thus, yielding pronounced savings.

\vspace{-1.5mm}
\section{Conclusion}

In this work we have proposed \shortname, an efficient and effective strategy to edit pre-trained models in the framework of task arithmetic. We started from the observation that the parameters showing the least variation in the fine-tuning process of a single task are also those minimally relevant for other tasks. Thus, we have leveraged them through a sparse learning process that promotes task localization and avoids task interference. 
A thorough experimental analysis across vision and language domains confirmed that \shortname yields state-of-the-art results in task addition and negation, showing a significant efficiency advantage over competitors. Moreover, with a tailored set of evaluations we assessed model linearization and function localization properties, providing insights on the inner functioning of our approach. 

Overall, we have discussed how preserving the regularities provided by a large scale pre-trained model are sufficient to maintain weight disentanglement and observe beneficial effects in task arithmetic. Future work may investigate whether explicitly enforcing localization constraints during fine-tuning could enhance performance and further advance model editing capabilities. 

\newpage
\section*{Reproducibility Statement}
We have made significant efforts to ensure the reproducibility of our results. Full implementation details are provided in Appendix \ref{sec:implementation}. Pseudocode for our algorithm is included in Appendix \ref{sec:mask_calibration} to clarify key steps, as well as practical design choices to address potential challenges in implementing our experiments. Additionally, we publicly released our code to further facilitate reproducibility at \url{https://github.com/iurada/talos-task-arithmetic}.

\section*{Acknowledgements}
The authors thank the reviewers and area chair for their valuable comments. M.C. also thanks Derek Tam and Colin Raffel for their fruitful discussions and feedback on the early state of this work. L.I. acknowledges the grant received from the European Union Next-GenerationEU (Piano Nazionale di Ripresa E Resilienza (PNRR)) DM 351 on Trustworthy AI. T.T. acknowledges the EU project ELSA - European Lighthouse on Secure and Safe AI.
This study was carried out within the FAIR - Future Artificial Intelligence Research and received funding from the European Union Next-GenerationEU (PIANO NAZIONALE DI RIPRESA E RESILIENZA (PNRR) – MISSIONE 4 COMPONENTE 2, INVESTIMENTO 1.3 – D.D. 1555 11/10/2022, PE00000013). 
This manuscript reflects only the authors’ views and opinions, neither the European Union nor the European Commission can be considered responsible for them. We acknowledge the CINECA award under the ISCRA initiative for the availability of high-performance computing resources and support.

\bibliography{iclr2025_conference}

\begin{thebibliography}{76}
\providecommand{\natexlab}[1]{#1}
\providecommand{\url}[1]{\texttt{#1}}
\expandafter\ifx\csname urlstyle\endcsname\relax
  \providecommand{\doi}[1]{doi: #1}\else
  \providecommand{\doi}{doi: \begingroup \urlstyle{rm}\Url}\fi

\bibitem[Amari(1996)]{amari1996neural}
Shun-ichi Amari.
\newblock Neural learning in structured parameter spaces-natural riemannian
  gradient.
\newblock In \emph{Advances in Neural Information Processing Systems
  (NeurIPS)}, 1996.
\newblock URL
  \url{https://proceedings.neurips.cc/paper_files/paper/1996/file/39e4973ba3321b80f37d9b55f63ed8b8-Paper.pdf}.

\bibitem[Ansell et~al.(2022)Ansell, Ponti, Korhonen, and
  Vuli{\'c}]{ansell-etal-2022-composable}
Alan Ansell, Edoardo Ponti, Anna Korhonen, and Ivan Vuli{\'c}.
\newblock Composable sparse fine-tuning for cross-lingual transfer.
\newblock In \emph{Proceedings of the 60th Annual Meeting of the Association
  for Computational Linguistics (Volume 1: Long Papers)}, 2022.
\newblock URL \url{https://aclanthology.org/2022.acl-long.125}.

\bibitem[Ansell et~al.(2024)Ansell, Vulić, Sterz, Korhonen, and
  Ponti]{ansell2024scaling}
Alan Ansell, Ivan Vulić, Hannah Sterz, Anna Korhonen, and Edoardo~M. Ponti.
\newblock Scaling sparse fine-tuning to large language models.
\newblock \emph{arXiv preprint arXiv:2401.16405}, 2024.
\newblock URL \url{https://arxiv.org/abs/2401.16405}.

\bibitem[Ben~Zaken et~al.(2022)Ben~Zaken, Goldberg, and
  Ravfogel]{ben-zaken-etal-2022-bitfit}
Elad Ben~Zaken, Yoav Goldberg, and Shauli Ravfogel.
\newblock {B}it{F}it: Simple parameter-efficient fine-tuning for
  transformer-based masked language-models.
\newblock In \emph{Proceedings of the 60th Annual Meeting of the Association
  for Computational Linguistics (Volume 2: Short Papers)}, 2022.
\newblock URL \url{https://aclanthology.org/2022.acl-short.1}.

\bibitem[Blalock et~al.(2020)Blalock, Gonzalez~Ortiz, Frankle, and
  Guttag]{blalock2020state}
Davis Blalock, Jose~Javier Gonzalez~Ortiz, Jonathan Frankle, and John Guttag.
\newblock What is the state of neural network pruning?
\newblock In \emph{Proceedings of Machine Learning and Systems (MLSys)}, 2020.
\newblock URL \url{https://arxiv.org/abs/2003.03033}.

\bibitem[Brown et~al.(2020)Brown, Mann, Ryder, Subbiah, Kaplan, Dhariwal,
  Neelakantan, Shyam, Sastry, Askell, Agarwal, Herbert-Voss, Krueger, Henighan,
  Child, Ramesh, Ziegler, Wu, Winter, Hesse, Chen, Sigler, Litwin, Gray, Chess,
  Clark, Berner, McCandlish, Radford, Sutskever, and Amodei]{brown2020gpt3}
Tom Brown, Benjamin Mann, Nick Ryder, Melanie Subbiah, Jared~D Kaplan, Prafulla
  Dhariwal, Arvind Neelakantan, Pranav Shyam, Girish Sastry, Amanda Askell,
  Sandhini Agarwal, Ariel Herbert-Voss, Gretchen Krueger, Tom Henighan, Rewon
  Child, Aditya Ramesh, Daniel Ziegler, Jeffrey Wu, Clemens Winter, Chris
  Hesse, Mark Chen, Eric Sigler, Mateusz Litwin, Scott Gray, Benjamin Chess,
  Jack Clark, Christopher Berner, Sam McCandlish, Alec Radford, Ilya Sutskever,
  and Dario Amodei.
\newblock Language models are few-shot learners.
\newblock In \emph{Advances in Neural Information Processing Systems
  (NeurIPS)}, 2020.
\newblock URL
  \url{https://proceedings.neurips.cc/paper_files/paper/2020/file/1457c0d6bfcb4967418bfb8ac142f64a-Paper.pdf}.

\bibitem[Chaudhry et~al.(2018)Chaudhry, Dokania, Ajanthan, and
  Torr]{chaudhry2018riemannian}
Arslan Chaudhry, Puneet~K Dokania, Thalaiyasingam Ajanthan, and Philip~HS Torr.
\newblock Riemannian walk for incremental learning: Understanding forgetting
  and intransigence.
\newblock In \emph{Proceedings of the European conference on computer vision
  (ECCV)}, 2018.
\newblock URL \url{https://arxiv.org/abs/1801.10112}.

\bibitem[Chen et~al.(2016)Chen, Xu, Zhang, and Guestrin]{chen2016training}
Tianqi Chen, Bing Xu, Chiyuan Zhang, and Carlos Guestrin.
\newblock Training deep nets with sublinear memory cost.
\newblock \emph{arXiv preprint arXiv:1604.06174}, 2016.
\newblock URL \url{https://arxiv.org/pdf/1604.06174}.

\bibitem[Cheng et~al.(2017)Cheng, Han, and Lu]{cheng2017resisc}
Gong Cheng, Junwei Han, and Xiaoqiang Lu.
\newblock Remote sensing image scene classification: Benchmark and state of the
  art.
\newblock \emph{Proceedings of the IEEE}, 105\penalty0 (10):\penalty0
  1865--1883, 2017.
\newblock URL \url{https://ieeexplore.ieee.org/document/7891544}.

\bibitem[Cimpoi et~al.(2014)Cimpoi, Maji, Kokkinos, Mohamed, and
  Vedaldi]{cimpoi2014dtd}
Mircea Cimpoi, Subhransu Maji, Iasonas Kokkinos, Sammy Mohamed, and Andrea
  Vedaldi.
\newblock Describing textures in the wild.
\newblock In \emph{Proceedings of the IEEE Conference on Computer Vision and
  Pattern Recognition (CVPR)}, 2014.
\newblock URL
  \url{https://openaccess.thecvf.com/content_cvpr_2014/papers/Cimpoi_Describing_Textures_in_2014_CVPR_paper.pdf}.

\bibitem[Dagan et~al.(2005)Dagan, Glickman, and Magnini]{dagan2005rte}
Ido Dagan, Oren Glickman, and Bernardo Magnini.
\newblock The pascal recognising textual entailment challenge.
\newblock In \emph{Machine learning challenges workshop}, 2005.
\newblock URL
  \url{https://citeseerx.ist.psu.edu/document?repid=rep1&type=pdf&doi=e808f28d411a958c5db81ceb111beb2638698f47}.

\bibitem[Davari \& Belilovsky(2024)Davari and Belilovsky]{davari2023model}
Mohammad~Reza Davari and Eugene Belilovsky.
\newblock Model breadcrumbs: Scaling multi-task model merging with sparse
  masks.
\newblock In \emph{European Conference on Computer Vision (ECCV)}, 2024.
\newblock URL \url{https://arxiv.org/abs/2312.06795}.

\bibitem[Deng et~al.(2009)Deng, Dong, Socher, Li, Li, and
  Fei-Fei]{deng2009imagenet}
Jia Deng, Wei Dong, Richard Socher, Li-Jia Li, Kai Li, and Li~Fei-Fei.
\newblock Imagenet: A large-scale hierarchical image database.
\newblock In \emph{IEEE conference on computer vision and pattern recognition
  (CVPR)}, 2009.
\newblock URL
  \url{https://projet.liris.cnrs.fr/imagine/pub/proceedings/CVPR-2009/data/papers/0103.pdf}.

\bibitem[Dettmers et~al.(2023)Dettmers, Pagnoni, Holtzman, and
  Zettlemoyer]{dettmers2024qlora}
Tim Dettmers, Artidoro Pagnoni, Ari Holtzman, and Luke Zettlemoyer.
\newblock Qlora: Efficient finetuning of quantized llms.
\newblock In \emph{Advances in Neural Information Processing Systems
  (NeurIPS)}, 2023.
\newblock URL \url{https://arxiv.org/abs/2305.14314}.

\bibitem[Dosovitskiy et~al.(2021)Dosovitskiy, Beyer, Kolesnikov, Weissenborn,
  Zhai, Unterthiner, Dehghani, Minderer, Heigold, Gelly,
  et~al.]{dosovitskiy2021vit}
Alexey Dosovitskiy, Lucas Beyer, Alexander Kolesnikov, Dirk Weissenborn,
  Xiaohua Zhai, Thomas Unterthiner, Mostafa Dehghani, Matthias Minderer, Georg
  Heigold, Sylvain Gelly, et~al.
\newblock An image is worth 16x16 words: Transformers for image recognition at
  scale.
\newblock In \emph{International Conference on Learning Representations
  (ICLR)}, 2021.
\newblock URL \url{https://arxiv.org/abs/2010.11929}.

\bibitem[Fisher(1922)]{fisher1922mathematical}
Ronald~A Fisher.
\newblock On the mathematical foundations of theoretical statistics.
\newblock \emph{Philosophical transactions of the Royal Society of London.
  Series A, containing papers of a mathematical or physical character},
  222\penalty0 (594-604):\penalty0 309--368, 1922.
\newblock URL
  \url{https://royalsocietypublishing.org/doi/10.1098/rsta.1922.0009}.

\bibitem[Frankle \& Carbin(2019)Frankle and Carbin]{frankle2018lottery}
Jonathan Frankle and Michael Carbin.
\newblock The lottery ticket hypothesis: Finding sparse, trainable neural
  networks.
\newblock In \emph{International Conference on Learning Representations
  (ICLR)}, 2019.
\newblock URL \url{https://arxiv.org/abs/1803.03635}.

\bibitem[Gadre et~al.(2024)Gadre, Ilharco, Fang, Hayase, Smyrnis, Nguyen,
  Marten, Wortsman, Ghosh, Zhang, et~al.]{gadre2024datacomp}
Samir~Yitzhak Gadre, Gabriel Ilharco, Alex Fang, Jonathan Hayase, Georgios
  Smyrnis, Thao Nguyen, Ryan Marten, Mitchell Wortsman, Dhruba Ghosh, Jieyu
  Zhang, et~al.
\newblock Datacomp: In search of the next generation of multimodal datasets.
\newblock In \emph{Advances in Neural Information Processing Systems
  (NeurIPS)}, 2024.
\newblock URL \url{https://arxiv.org/abs/2304.14108}.

\bibitem[Guo et~al.(2021)Guo, Rush, and Kim]{guo-etal-2021-parameter}
Demi Guo, Alexander Rush, and Yoon Kim.
\newblock Parameter-efficient transfer learning with diff pruning.
\newblock In \emph{Proceedings of the 59th Annual Meeting of the Association
  for Computational Linguistics and the 11th International Joint Conference on
  Natural Language Processing (Volume 1: Long Papers)}, 2021.
\newblock URL \url{https://aclanthology.org/2021.acl-long.378}.

\bibitem[Havasi et~al.(2020)Havasi, Jenatton, Fort, Liu, Snoek,
  Lakshminarayanan, Dai, and Tran]{havasi2020training}
Marton Havasi, Rodolphe Jenatton, Stanislav Fort, Jeremiah~Zhe Liu, Jasper
  Snoek, Balaji Lakshminarayanan, Andrew~M Dai, and Dustin Tran.
\newblock Training independent subnetworks for robust prediction.
\newblock \emph{arXiv preprint arXiv:2010.06610}, 2020.
\newblock URL \url{https://arxiv.org/abs/2010.06610}.

\bibitem[Helber et~al.(2019)Helber, Bischke, Dengel, and
  Borth]{helber2019eurosat}
Patrick Helber, Benjamin Bischke, Andreas Dengel, and Damian Borth.
\newblock Eurosat: A novel dataset and deep learning benchmark for land use and
  land cover classification.
\newblock \emph{IEEE Journal of Selected Topics in Applied Earth Observations
  and Remote Sensing}, 12\penalty0 (7):\penalty0 2217--2226, 2019.
\newblock URL \url{https://ieeexplore.ieee.org/document/8736785}.

\bibitem[Hinton(2022)]{hinton2022forward}
Geoffrey Hinton.
\newblock The forward-forward algorithm: Some preliminary investigations.
\newblock \emph{arXiv preprint arXiv:2212.13345}, 2022.
\newblock URL \url{https://arxiv.org/pdf/2212.13345}.

\bibitem[Houlsby et~al.(2019)Houlsby, Giurgiu, Jastrzebski, Morrone,
  De~Laroussilhe, Gesmundo, Attariyan, and Gelly]{pmlr-v97-houlsby19a}
Neil Houlsby, Andrei Giurgiu, Stanislaw Jastrzebski, Bruna Morrone, Quentin
  De~Laroussilhe, Andrea Gesmundo, Mona Attariyan, and Sylvain Gelly.
\newblock Parameter-efficient transfer learning for {NLP}.
\newblock In \emph{International Conference on Machine Learning (ICML)}, 2019.
\newblock URL \url{https://arxiv.org/abs/1902.00751}.

\bibitem[Hu et~al.(2022)Hu, yelong shen, Wallis, Allen-Zhu, Li, Wang, Wang, and
  Chen]{hu2022lora}
Edward~J Hu, yelong shen, Phillip Wallis, Zeyuan Allen-Zhu, Yuanzhi Li, Shean
  Wang, Lu~Wang, and Weizhu Chen.
\newblock Lo{RA}: Low-rank adaptation of large language models.
\newblock In \emph{International Conference on Learning Representations
  (ICLR)}, 2022.
\newblock URL \url{https://arxiv.org/abs/2106.09685}.

\bibitem[Ilharco et~al.(2022)Ilharco, Wortsman, Gadre, Song, Hajishirzi,
  Kornblith, Farhadi, and Schmidt]{ilharco2022patching}
Gabriel Ilharco, Mitchell Wortsman, Samir~Yitzhak Gadre, Shuran Song, Hannaneh
  Hajishirzi, Simon Kornblith, Ali Farhadi, and Ludwig Schmidt.
\newblock Patching open-vocabulary models by interpolating weights.
\newblock In \emph{Advances in Neural Information Processing Systems
  (NeurIPS)}, 2022.
\newblock URL \url{https://arxiv.org/abs/2208.05592}.

\bibitem[Ilharco et~al.(2023)Ilharco, Ribeiro, Wortsman, Gururangan, Schmidt,
  Hajishirzi, and Farhadi]{ilharco2023editing}
Gabriel Ilharco, Marco~Tulio Ribeiro, Mitchell Wortsman, Suchin Gururangan,
  Ludwig Schmidt, Hannaneh Hajishirzi, and Ali Farhadi.
\newblock Editing models with task arithmetic.
\newblock In \emph{International Conference on Learning Representations
  (ICLR)}, 2023.
\newblock URL \url{https://arxiv.org/abs/2110.08207}.

\bibitem[Jin et~al.(2023)Jin, Ren, Preotiuc-Pietro, and Cheng]{jin2023dataless}
Xisen Jin, Xiang Ren, Daniel Preotiuc-Pietro, and Pengxiang Cheng.
\newblock Dataless knowledge fusion by merging weights of language models.
\newblock In \emph{International Conference on Learning Representations
  (ICLR)}, 2023.
\newblock URL \url{https://arxiv.org/abs/2212.09849}.

\bibitem[Khot et~al.(2020)Khot, Clark, Guerquin, Jansen, and
  Sabharwal]{khot2020qasc}
Tushar Khot, Peter Clark, Michal Guerquin, Peter Jansen, and Ashish Sabharwal.
\newblock Qasc: A dataset for question answering via sentence composition.
\newblock In \emph{Proceedings of the AAAI Conference on Artificial
  Intelligence (AAAI)}, 2020.
\newblock URL \url{https://arxiv.org/abs/1910.11473}.

\bibitem[Krause et~al.(2013)Krause, Stark, Deng, and Fei-Fei]{krause2013cars}
Jonathan Krause, Michael Stark, Jia Deng, and Li~Fei-Fei.
\newblock 3d object representations for fine-grained categorization.
\newblock In \emph{Proceedings of the IEEE International Conference on Computer
  Vision (ICCV) Workshops}, 2013.
\newblock URL
  \url{https://www.cv-foundation.org/openaccess/content_iccv_workshops_2013/W19/papers/Krause_3D_Object_Representations_2013_ICCV_paper.pdf}.

\bibitem[Kunstner et~al.(2019)Kunstner, Hennig, and
  Balles]{kunstner2019limitations}
Frederik Kunstner, Philipp Hennig, and Lukas Balles.
\newblock Limitations of the empirical fisher approximation for natural
  gradient descent.
\newblock In \emph{Advances in Neural Information Processing Systems
  (NeurIPS)}, 2019.
\newblock URL \url{https://arxiv.org/abs/1905.12558}.

\bibitem[Kwon et~al.(2022)Kwon, Kim, Mahoney, Hassoun, Keutzer, and
  Gholami]{kwon2022fast}
Woosuk Kwon, Sehoon Kim, Michael~W Mahoney, Joseph Hassoun, Kurt Keutzer, and
  Amir Gholami.
\newblock A fast post-training pruning framework for transformers.
\newblock In \emph{Advances in Neural Information Processing Systems
  (NeurIPS)}, 2022.
\newblock URL
  \url{https://proceedings.neurips.cc/paper_files/paper/2022/file/987bed997ab668f91c822a09bce3ea12-Paper-Conference.pdf}.

\bibitem[LeCun(1998)]{lecun1998mnist}
Yann LeCun.
\newblock The mnist database of handwritten digits, 1998.
\newblock URL \url{http://yann.lecun.com/exdb/mnist/}.

\bibitem[Levesque et~al.(2012)Levesque, Davis, and
  Morgenstern]{levesque2012wsc}
Hector Levesque, Ernest Davis, and Leora Morgenstern.
\newblock The winograd schema challenge.
\newblock In \emph{Thirteenth international conference on the principles of
  knowledge representation and reasoning (KR)}, 2012.
\newblock URL \url{https://cdn.aaai.org/ocs/4492/4492-21843-1-PB.pdf}.

\bibitem[Li \& Liang(2021)Li and Liang]{prefix2021}
Xiang~Lisa Li and Percy Liang.
\newblock Prefix-tuning: Optimizing continuous prompts for generation.
\newblock \emph{arXiv preprint arXiv:2101.00190}, 2021.
\newblock URL \url{https://arxiv.org/abs/2101.00190}.

\bibitem[Liao et~al.(2023)Liao, Meng, and Monz]{liao-etal-2023-parameter}
Baohao Liao, Yan Meng, and Christof Monz.
\newblock Parameter-efficient fine-tuning without introducing new latency.
\newblock In \emph{Proceedings of the 61st Annual Meeting of the Association
  for Computational Linguistics (Volume 1: Long Papers)}, 2023.
\newblock URL \url{https://aclanthology.org/2023.acl-long.233}.

\bibitem[Liu et~al.(2022)Liu, Tam, Muqeeth, Mohta, Huang, Bansal, and
  Raffel]{liu2022few}
Haokun Liu, Derek Tam, Mohammed Muqeeth, Jay Mohta, Tenghao Huang, Mohit
  Bansal, and Colin~A Raffel.
\newblock Few-shot parameter-efficient fine-tuning is better and cheaper than
  in-context learning.
\newblock In \emph{Advances in Neural Information Processing Systems
  (NeurIPS)}, 2022.
\newblock URL \url{https://arxiv.org/abs/2205.05638}.

\bibitem[Liu et~al.(2024)Liu, Wang, Yin, Molchanov, Wang, Cheng, and
  Chen]{liu24dora}
Shih-Yang Liu, Chien-Yi Wang, Hongxu Yin, Pavlo Molchanov, Yu-Chiang~Frank
  Wang, Kwang-Ting Cheng, and Min-Hung Chen.
\newblock {D}o{RA}: Weight-decomposed low-rank adaptation.
\newblock In \emph{International Conference on Machine Learning (ICML)}, 2024.
\newblock URL \url{https://arxiv.org/abs/2402.09353}.

\bibitem[Loshchilov \& Hutter(2019)Loshchilov and Hutter]{loshchilov2019adamw}
Ilya Loshchilov and Frank Hutter.
\newblock Decoupled weight decay regularization.
\newblock In \emph{International Conference on Learning Representations
  (ICLR)}, 2019.
\newblock URL \url{https://openreview.net/forum?id=Bkg6RiCqY7}.

\bibitem[Malladi et~al.(2023{\natexlab{a}})Malladi, Gao, Nichani, Damian, Lee,
  Chen, and Arora]{malladi2023fine}
Sadhika Malladi, Tianyu Gao, Eshaan Nichani, Alex Damian, Jason~D Lee, Danqi
  Chen, and Sanjeev Arora.
\newblock Fine-tuning language models with just forward passes.
\newblock In \emph{Advances in Neural Information Processing Systems
  (NeurIPS)}, 2023{\natexlab{a}}.
\newblock URL
  \url{https://proceedings.neurips.cc/paper_files/paper/2023/file/a627810151be4d13f907ac898ff7e948-Paper-Conference.pdf}.

\bibitem[Malladi et~al.(2023{\natexlab{b}})Malladi, Wettig, Yu, Chen, and
  Arora]{malladi2023kernel}
Sadhika Malladi, Alexander Wettig, Dingli Yu, Danqi Chen, and Sanjeev Arora.
\newblock A kernel-based view of language model fine-tuning.
\newblock In \emph{International Conference on Machine Learning (ICML)},
  2023{\natexlab{b}}.
\newblock URL
  \url{https://proceedings.mlr.press/v202/malladi23a/malladi23a.pdf}.

\bibitem[Mallya \& Lazebnik(2018)Mallya and Lazebnik]{mallya2018packnet}
Arun Mallya and Svetlana Lazebnik.
\newblock Packnet: Adding multiple tasks to a single network by iterative
  pruning.
\newblock In \emph{Proceedings of the IEEE conference on Computer Vision and
  Pattern Recognition (CVPR)}, 2018.
\newblock URL \url{https://arxiv.org/abs/1711.05769}.

\bibitem[Mallya et~al.(2018)Mallya, Davis, and Lazebnik]{mallya2018piggyback}
Arun Mallya, Dillon Davis, and Svetlana Lazebnik.
\newblock Piggyback: Adapting a single network to multiple tasks by learning to
  mask weights.
\newblock In \emph{Proceedings of the European conference on computer vision
  (ECCV)}, 2018.
\newblock URL \url{https://arxiv.org/abs/1801.06519}.

\bibitem[Mangrulkar et~al.(2022)Mangrulkar, Gugger, Debut, Belkada, Paul, and
  Bossan]{peftLib}
Sourab Mangrulkar, Sylvain Gugger, Lysandre Debut, Younes Belkada, Sayak Paul,
  and Benjamin Bossan.
\newblock Peft: State-of-the-art parameter-efficient fine-tuning methods.
\newblock \url{https://github.com/huggingface/peft}, 2022.

\bibitem[Matena \& Raffel(2022)Matena and Raffel]{mergingfisher_2024}
Michael Matena and Colin Raffel.
\newblock Merging models with fisher-weighted averaging.
\newblock In \emph{Advances in Neural Information Processing Systems
  (NeurIPS)}, 2022.
\newblock URL \url{https://arxiv.org/abs/2111.09832}.

\bibitem[Netzer et~al.(2011)Netzer, Wang, Coates, Bissacco, Wu, Ng,
  et~al.]{netzer2011svhn}
Yuval Netzer, Tao Wang, Adam Coates, Alessandro Bissacco, Baolin Wu, Andrew~Y
  Ng, et~al.
\newblock Reading digits in natural images with unsupervised feature learning.
\newblock In \emph{Advances in Neural Information Processing Systems (NeurIPS)
  Workshops}, 2011.
\newblock URL
  \url{https://static.googleusercontent.com/media/research.google.com/it//pubs/archive/37648.pdf}.

\bibitem[Ortiz-Jim{\'e}nez et~al.(2021)Ortiz-Jim{\'e}nez, Moosavi-Dezfooli, and
  Frossard]{ortiz2021can}
Guillermo Ortiz-Jim{\'e}nez, Seyed-Mohsen Moosavi-Dezfooli, and Pascal
  Frossard.
\newblock What can linearized neural networks actually say about
  generalization?
\newblock In \emph{Advances in Neural Information Processing Systems
  (NeurIPS)}, 2021.
\newblock URL
  \url{https://proceedings.neurips.cc/paper/2021/file/4b5deb9a14d66ab0acc3b8a2360cde7c-Paper.pdf}.

\bibitem[Ortiz-Jimenez et~al.(2023)Ortiz-Jimenez, Favero, and
  Frossard]{Tangent_task_arith_2023}
Guillermo Ortiz-Jimenez, Alessandro Favero, and Pascal Frossard.
\newblock Task arithmetic in the tangent space: Improved editing of pre-trained
  models.
\newblock In \emph{Advances in Neural Information Processing Systems
  (NeurIPS)}, 2023.
\newblock URL \url{https://openreview.net/pdf?id=0A9f2jZDGW}.

\bibitem[Panda et~al.(2024)Panda, Isik, Qi, Koyejo, Weissman, and
  Mittal]{panda2024lottery}
Ashwinee Panda, Berivan Isik, Xiangyu Qi, Sanmi Koyejo, Tsachy Weissman, and
  Prateek Mittal.
\newblock Lottery ticket adaptation: Mitigating destructive interference in
  llms.
\newblock \emph{arXiv preprint arXiv:2406.16797}, 2024.
\newblock URL \url{https://arxiv.org/abs/2406.16797}.

\bibitem[Pascanu \& Bengio(2013)Pascanu and Bengio]{pascanu2013revisiting}
R~Pascanu and Y~Bengio.
\newblock Revisiting natural gradient for deep networks.
\newblock \emph{arXiv preprint arXiv:1301.3584}, 2013.
\newblock URL \url{https://arxiv.org/abs/1301.3584}.

\bibitem[Pennington \& Worah(2018)Pennington and Worah]{pennington2018spectrum}
Jeffrey Pennington and Pratik Worah.
\newblock The spectrum of the fisher information matrix of a
  single-hidden-layer neural network.
\newblock In \emph{Advances in Neural Information Processing Systems
  (NeurIPS)}, 2018.
\newblock URL
  \url{https://papers.nips.cc/paper_files/paper/2018/file/18bb68e2b38e4a8ce7cf4f6b2625768c-Paper.pdf}.

\bibitem[Pfeiffer et~al.(2020)Pfeiffer, R\"uckl\'{e}, Poth, Kamath, Vuli\'{c},
  Ruder, Cho, and Gurevych]{pfeiffer2020AdapterHub}
Jonas Pfeiffer, Andreas R\"uckl\'{e}, Clifton Poth, Aishwarya Kamath, Ivan
  Vuli\'{c}, Sebastian Ruder, Kyunghyun Cho, and Iryna Gurevych.
\newblock Adapterhub: A framework for adapting transformers.
\newblock In \emph{Proceedings of the 2020 Conference on Empirical Methods in
  Natural Language Processing (EMNLP): Systems Demonstrations}, 2020.
\newblock URL \url{https://www.aclweb.org/anthology/2020.emnlp-demos.7}.

\bibitem[Poth et~al.(2023)Poth, Sterz, Paul, Purkayastha, Engl{\"a}nder, Imhof,
  Vuli{\'c}, Ruder, Gurevych, and Pfeiffer]{poth-etal-2023-adapters}
Clifton Poth, Hannah Sterz, Indraneil Paul, Sukannya Purkayastha, Leon
  Engl{\"a}nder, Timo Imhof, Ivan Vuli{\'c}, Sebastian Ruder, Iryna Gurevych,
  and Jonas Pfeiffer.
\newblock Adapters: A unified library for parameter-efficient and modular
  transfer learning.
\newblock In \emph{Proceedings of the 2023 Conference on Empirical Methods in
  Natural Language Processing (EMNLP): System Demonstrations}, 2023.
\newblock URL \url{https://aclanthology.org/2023.emnlp-demo.13}.

\bibitem[Radford et~al.(2021)Radford, Kim, Hallacy, Ramesh, Goh, Agarwal,
  Sastry, Askell, Mishkin, Clark, et~al.]{radford2021clip}
Alec Radford, Jong~Wook Kim, Chris Hallacy, Aditya Ramesh, Gabriel Goh,
  Sandhini Agarwal, Girish Sastry, Amanda Askell, Pamela Mishkin, Jack Clark,
  et~al.
\newblock Learning transferable visual models from natural language
  supervision.
\newblock In \emph{International Conference on Machine Learning (ICML)}, 2021.
\newblock URL \url{https://arxiv.org/abs/2103.00020}.

\bibitem[Raffel(2023)]{raffel2023building}
Colin Raffel.
\newblock Building machine learning models like open source software.
\newblock \emph{Communications of the ACM}, 66\penalty0 (2):\penalty0 38--40,
  2023.
\newblock URL \url{https://dl.acm.org/doi/pdf/10.1145/3545111}.

\bibitem[Raffel et~al.(2020)Raffel, Shazeer, Roberts, Lee, Narang, Matena,
  Zhou, Li, and Liu]{2020t5}
Colin Raffel, Noam Shazeer, Adam Roberts, Katherine Lee, Sharan Narang, Michael
  Matena, Yanqi Zhou, Wei Li, and Peter~J. Liu.
\newblock Exploring the limits of transfer learning with a unified text-to-text
  transformer.
\newblock \emph{Journal of Machine Learning Research}, 21\penalty0
  (140):\penalty0 1--67, 2020.
\newblock URL \url{http://jmlr.org/papers/v21/20-074.html}.

\bibitem[Ram{\'e} et~al.(2023)Ram{\'e}, Ahuja, Zhang, Cord, Bottou, and
  Lopez-Paz]{rame2023model}
Alexandre Ram{\'e}, Kartik Ahuja, Jianyu Zhang, Matthieu Cord, L{\'e}on Bottou,
  and David Lopez-Paz.
\newblock Model ratatouille: Recycling diverse models for out-of-distribution
  generalization.
\newblock In \emph{International Conference on Machine Learning (ICML)}, 2023.
\newblock URL \url{https://arxiv.org/abs/2212.10445}.

\bibitem[Sakaguchi et~al.(2021)Sakaguchi, Bras, Bhagavatula, and
  Choi]{sakaguchi2021winogrande}
Keisuke Sakaguchi, Ronan~Le Bras, Chandra Bhagavatula, and Yejin Choi.
\newblock Winogrande: An adversarial winograd schema challenge at scale.
\newblock \emph{Communications of the ACM}, 64\penalty0 (9):\penalty0 99--106,
  2021.
\newblock URL \url{https://dl.acm.org/doi/pdf/10.1145/3474381}.

\bibitem[Sharma et~al.(2018)Sharma, Allen, Bakhshandeh, and
  Mostafazadeh]{sharma2018storycloze}
Rishi Sharma, James Allen, Omid Bakhshandeh, and Nasrin Mostafazadeh.
\newblock Tackling the story ending biases in the story cloze test.
\newblock In \emph{Proceedings of the 56th Annual Meeting of the Association
  for Computational Linguistics (ACL)}, 2018.
\newblock URL \url{https://aclanthology.org/P18-2119.pdf}.

\bibitem[Stallkamp et~al.(2011)Stallkamp, Schlipsing, Salmen, and
  Igel]{stallkamp2011gtsrb}
Johannes Stallkamp, Marc Schlipsing, Jan Salmen, and Christian Igel.
\newblock The german traffic sign recognition benchmark: a multi-class
  classification competition.
\newblock In \emph{International Joint Conference on Neural Networks (IJCNN)},
  2011.
\newblock URL \url{https://ieeexplore.ieee.org/document/6033395}.

\bibitem[Sung et~al.(2021)Sung, Nair, and Raffel]{sung2021training}
Yi-Lin Sung, Varun Nair, and Colin~A Raffel.
\newblock Training neural networks with fixed sparse masks.
\newblock In \emph{Advances in Neural Information Processing Systems
  (NeurIPS)}, 2021.
\newblock URL \url{https://arxiv.org/abs/2111.09839}.

\bibitem[Sung et~al.(2024)Sung, Yoon, and Bansal]{sung2023ecoflap}
Yi-Lin Sung, Jaehong Yoon, and Mohit Bansal.
\newblock Ecoflap: Efficient coarse-to-fine layer-wise pruning for
  vision-language models.
\newblock In \emph{International Conference on Learning Representations
  (ICLR)}, 2024.
\newblock URL \url{https://arxiv.org/pdf/2310.02998}.

\bibitem[Tafjord et~al.(2019)Tafjord, Gardner, Lin, and
  Clark]{tafjord2019quartz}
Oyvind Tafjord, Matt Gardner, Kevin Lin, and Peter Clark.
\newblock Quartz: An open-domain dataset of qualitative relationship questions.
\newblock In \emph{Proceedings of the 2019 conference on empirical methods in
  natural language processing (EMNLP)}, 2019.
\newblock URL \url{https://arxiv.org/abs/1909.03553}.

\bibitem[Tanaka et~al.(2020)Tanaka, Kunin, Yamins, and
  Ganguli]{tanaka2020synflow}
Hidenori Tanaka, Daniel Kunin, Daniel~L Yamins, and Surya Ganguli.
\newblock Pruning neural networks without any data by iteratively conserving
  synaptic flow.
\newblock In \emph{Advances in Neural Information Processing Systems
  (NeurIPS)}, 2020.
\newblock URL \url{https://arxiv.org/abs/2006.05467}.

\bibitem[Tang et~al.(2024)Tang, Shen, Luo, Zhan, Hu, Du, Chen, and
  Tao]{tang2023parameter}
Anke Tang, Li~Shen, Yong Luo, Yibing Zhan, Han Hu, Bo~Du, Yixin Chen, and
  Dacheng Tao.
\newblock Parameter efficient multi-task model fusion with partial
  linearization.
\newblock In \emph{International Conference on Learning Representations
  (ICLR)}, 2024.
\newblock URL \url{https://arxiv.org/abs/2310.04742}.

\bibitem[Wang et~al.(2024)Wang, Dimitriadis, Ortiz-Jimenez, Fleuret, and
  Frossard]{wang2024consensus}
Ke~Wang, Nikolaos Dimitriadis, Guillermo Ortiz-Jimenez, Fran{\c{c}}ois Fleuret,
  and Pascal Frossard.
\newblock Localizing task information for improved model merging and
  compression.
\newblock In \emph{International Conference on Machine Learning (ICML)}, 2024.
\newblock URL \url{https://arxiv.org/abs/2405.07813}.

\bibitem[Wang et~al.(2023)Wang, Li, and Sun]{wang2023ntksap}
Yite Wang, Dawei Li, and Ruoyu Sun.
\newblock Ntk-sap: Improving neural network pruning by aligning training
  dynamics.
\newblock In \emph{International Conference on Learning Representations
  (ICLR)}, 2023.
\newblock URL \url{https://arxiv.org/abs/2304.02840}.

\bibitem[Wortsman et~al.(2020)Wortsman, Ramanujan, Liu, Kembhavi, Rastegari,
  Yosinski, and Farhadi]{wortsman2020supermasks}
Mitchell Wortsman, Vivek Ramanujan, Rosanne Liu, Aniruddha Kembhavi, Mohammad
  Rastegari, Jason Yosinski, and Ali Farhadi.
\newblock Supermasks in superposition.
\newblock In \emph{Advances in Neural Information Processing Systems
  (NeurIPS)}, 2020.
\newblock URL \url{https://arxiv.org/abs/2006.14769}.

\bibitem[Wortsman et~al.(2022)Wortsman, Ilharco, Gadre, Roelofs, Gontijo-Lopes,
  Morcos, Namkoong, Farhadi, Carmon, Kornblith, et~al.]{wortsman2022model}
Mitchell Wortsman, Gabriel Ilharco, Samir~Ya Gadre, Rebecca Roelofs, Raphael
  Gontijo-Lopes, Ari~S Morcos, Hongseok Namkoong, Ali Farhadi, Yair Carmon,
  Simon Kornblith, et~al.
\newblock Model soups: averaging weights of multiple fine-tuned models improves
  accuracy without increasing inference time.
\newblock In \emph{International Conference on Machine Learning (ICML)}, 2022.
\newblock URL \url{https://arxiv.org/abs/2203.05482}.

\bibitem[Xiao et~al.(2016)Xiao, Ehinger, Hays, Torralba, and
  Oliva]{xiao2016sun}
Jianxiong Xiao, Krista~A Ehinger, James Hays, Antonio Torralba, and Aude Oliva.
\newblock Sun database: Exploring a large collection of scene categories.
\newblock \emph{International Journal of Computer Vision}, 119:\penalty0 3--22,
  2016.
\newblock URL
  \url{https://link.springer.com/article/10.1007/s11263-014-0748-y}.

\bibitem[Xu et~al.(2021)Xu, Luo, Zhang, Tan, Chang, Huang, and
  Huang]{xu-etal-2021-raise}
Runxin Xu, Fuli Luo, Zhiyuan Zhang, Chuanqi Tan, Baobao Chang, Songfang Huang,
  and Fei Huang.
\newblock Raise a child in large language model: Towards effective and
  generalizable fine-tuning.
\newblock In \emph{Proceedings of the 2021 Conference on Empirical Methods in
  Natural Language Processing (EMNLP)}, 2021.
\newblock URL \url{https://aclanthology.org/2021.emnlp-main.749}.

\bibitem[Xu et~al.(2020)Xu, Wang, Wang, O'Neill, and Zhu]{xu2020one}
Shichao Xu, Yixuan Wang, Yanzhi Wang, Zheng O'Neill, and Qi~Zhu.
\newblock One for many: Transfer learning for building hvac control.
\newblock In \emph{International Conference on Systems for Energy-Efficient
  Built Environments (BuildSys)}, 2020.
\newblock URL \url{https://arxiv.org/abs/2008.03625}.

\bibitem[Yadav et~al.(2023)Yadav, Tam, Choshen, Raffel, and
  Bansal]{yadav2023ties}
Prateek Yadav, Derek Tam, Leshem Choshen, Colin Raffel, and Mohit Bansal.
\newblock Ties-merging: Resolving interference when merging models.
\newblock In \emph{Advances in Neural Information Processing Systems
  (NeurIPS)}, 2023.
\newblock URL \url{https://arxiv.org/abs/2306.01708}.

\bibitem[Yang et~al.(2024)Yang, Wang, Shen, Liu, Guo, Wang, and
  Tao]{AdaMerging_ICLR_2024}
Enneng Yang, Zhenyi Wang, Li~Shen, Shiwei Liu, Guibing Guo, Xingwei Wang, and
  Dacheng Tao.
\newblock Adamerging: Adaptive model merging for multi-task learning.
\newblock In \emph{International Conference on Learning Representations
  (ICLR)}, 2024.
\newblock URL \url{https://arxiv.org/abs/2310.02575}.

\bibitem[Yang et~al.(2015)Yang, Yih, and Meek]{yang2015wikiqa}
Yi~Yang, Wen-tau Yih, and Christopher Meek.
\newblock Wikiqa: A challenge dataset for open-domain question answering.
\newblock In \emph{Proceedings of the 2015 conference on empirical methods in
  natural language processing (EMNLP)}, 2015.
\newblock URL \url{https://aclanthology.org/D15-1237.pdf}.

\bibitem[Yu et~al.(2024)Yu, Yu, Yu, Huang, and Li]{dareSupermario}
Le~Yu, Bowen Yu, Haiyang Yu, Fei Huang, and Yongbin Li.
\newblock Language models are super mario: Absorbing abilities from homologous
  models as a free lunch.
\newblock In \emph{International Conference on Machine Learning (ICML)}, 2024.
\newblock URL \url{https://arxiv.org/abs/2311.03099}.

\bibitem[Zhang et~al.(2019)Zhang, Baldridge, and He]{zhang2019paws}
Yuan Zhang, Jason Baldridge, and Luheng He.
\newblock Paws: Paraphrase adversaries from word scrambling.
\newblock In \emph{2024 Annual Conference of the North American Chapter of the
  Association for Computational Linguistics (NAACL)}, 2019.
\newblock URL \url{https://arxiv.org/abs/1904.01130}.

\end{thebibliography}
\bibliographystyle{iclr2025_conference}

\appendix
\newpage
\section{Appendix}

\subsection{Implementation Details}\label{sec:implementation}

\textbf{Computational resources.} We execute all the vision experiments using ViT-B/32, ViT-B/16, and ViT-L/14 on a machine equipped with two NVIDIA GeForce RTX 2080 Ti (11 GB VRAM), an Intel Core i7-9800X CPU @ 3.80GHz and 64 GB of RAM. For all the language experiments using T5-Small, T5-Base, and T5-Large we employ a machine equipped with a a single NVIDIA A100 SXM (64 GB VRAM), an Intel Xeon Platinum 8358 CPU @ 2.60GHz and 64 GB of RAM.

\textbf{Starter code.} We developed our codebase starting from the repositories provided by \citet{Tangent_task_arith_2023}\footnote{\url{https://github.com/gortizji/tangent_task_arithmetic}} (based on the code by \citet{ilharco2022patching, ilharco2023editing}\footnote{\url{https://github.com/mlfoundations/task_vectors}}) and \cite{yadav2023ties}\footref{ties_code}, which allow to reproduce the full fine-tuning results (Non-linear FT and Linearized FT). TIES-Merging \citep{yadav2023ties}\footnote{\url{https://github.com/prateeky2806/ties-merging}\label{ties_code}}, TALL Mask / Consensus \citep{wang2024consensus}\footnote{\url{https://github.com/nik-dim/tall_masks}}, DARE \citep{dareSupermario}\footnote{\url{https://github.com/yule-BUAA/MergeLM}}, Breadcrumbs \citep{davari2023model}\footnote{\url{https://github.com/rezazzr/breadcrumbs}} and LoTA \citep{panda2024lottery}\footnote{\url{https://github.com/kiddyboots216/lottery-ticket-adaptation}} provide official implementations of their methods from which we carefully adapted their code to work within the Task Arithmetic framework. L-LoRA \citep{tang2023parameter} unfortunately doesn't provide any official implementation, but the guidelines in the paper are sufficient to reproduce their results. To this end, we used the \texttt{peft} library \citep{peftLib}\footnote{\url{https://github.com/huggingface/peft}} for implementing the LoRA modules.

\textbf{Hyperparameter selection.} As highlighted by \citet{Tangent_task_arith_2023}, task vectors that perform well in Task Negation tend to exhibit higher degrees of weight disentanglement in Task Addition. This relationship informed our hyperparameter selection strategy. For each method, we cross-validate its hyperparameters on each individual task by leveraging Task Negation performance on a small held-out portion of the training set, as implemented by \citet{ilharco2023editing, Tangent_task_arith_2023}. It's important to note that hyperparameter selection shall not be performed separately for addition and negation, as each choice of hyperparameters yields a unique task vector. 
Hyperparameter search of each method is carried out according to the guidelines presented in each paper. Specifically, for \textbf{post-hoc} methods, the sparsity ratio is searched in the set $k \in \{0.1, 0.2, ..., 0.9, 0.95, 0.99\}$. Furthermore, for TALL Mask / Consensus \citep{wang2024consensus} we also tune the \emph{consensus threshold} in the set $\{0,...,T\}$, where $T$ is the number of tasks. For Breadcrumbs \citep{davari2023model} we also tune the percentage of top-$k$ parameters considered outliers, using values from the set $\{0.8, 0.9, 0.95, 0.99, 0.992, 0.994, ..., 0.999\}$.
Regarding \textbf{parameter-efficient fine-tuning} methods, when using L-LoRA \citep{tang2023parameter} we progressively reduce its rank $r \in \{512, 256, 128, 64, 32, 16, 8\}$. While, for LoTA \citep{panda2024lottery} and our method we tune sparsity at the task level using values in the set $\{0.1, 0.2, ..., 0.9, 0.95, 0.99\}$. Regarding the amount of data used to perform mask calibration on each task, we align with \citet{panda2024lottery} by using the validation split as it accounts for the 10\% of the total training data. For LoTA, we set the number of iterations for mask calibration so to match the number of mask calibration rounds used by our method (further details at Section \ref{sec:mask_calibration}). This ensures that the drop in performance is negligible with respect to using the full training split while significantly reducing the computational overhead.

\textbf{Datasets \& Tasks.} In line with what introduced in \citet{ilharco2022patching, ilharco2023editing, Tangent_task_arith_2023}, our vision experiments consider %center on 
image classification across various domains.
We adhere to the proposed experimental setup by utilizing eight datasets: Cars \citep{krause2013cars}, DTD \citep{cimpoi2014dtd}, EuroSAT \citep{helber2019eurosat}, GTSRB \citep{stallkamp2011gtsrb}, MNIST \citep{lecun1998mnist}, RESISC45 \citep{cheng2017resisc}, SUN397 \citep{xiao2016sun} and SVHN \citep{netzer2011svhn}. 

For the natural language processing (NLP) experiments, we follow the methodology outlined in \citet{yadav2023ties}, incorporating seven prescribed datasets: three regarding question answering (QASC \citep{khot2020qasc}, WikiQA \citep{yang2015wikiqa} and QuaRTz \citep{tafjord2019quartz}), one for paraphrase identification (PAWS \citep{zhang2019paws}), one focusing on sentence completion (Story Cloze \citep{sharma2018storycloze}) and two for coreference resolution (Winogrande \citep{sakaguchi2021winogrande} and WSC \citep{levesque2012wsc}). 
Concerning Task Negation, we align with \citet{Tangent_task_arith_2023} and consider ImageNet \citep{deng2009imagenet} as the control dataset for vision experiments, while for NLP, we utilize RTE \citep{dagan2005rte}, as it provides a distinct task (\ie natural language inference) with respect to the others considered for the NLP experiments.

\textbf{Architectures \& Pre-trained models.} 
By following \citet{ilharco2023editing, Tangent_task_arith_2023, yadav2023ties},  
on vision experiments, we use three variants of CLIP \citep{radford2021clip} with ViT-B/32, ViT-B/16, and ViT-L/14 models \citep{dosovitskiy2021vit}. Regarding the NLP experiments, we employ T5-Small, T5-Base, and T5-Large models \citep{2020t5}.

\textbf{Fine-tuning details.} All fine-tuning experiments on vision adhere to the training protocol outlined by \cite{ilharco2022patching, ilharco2023editing, Tangent_task_arith_2023}, with minor modifications made to the training code to accommodate the additional baselines and our method. Specifically, we fine-tune all datasets starting from the same CLIP pre-trained checkpoint, which is obtained from the \texttt{open\_clip} repository \citep{gadre2024datacomp}. Each model is fine-tuned for 2,000 iterations with a batch size of 128, a learning rate of $10^{-5}$, and a cosine annealing learning rate schedule with 200 warm-up steps. We use the AdamW optimizer \citep{loshchilov2019adamw}. Following \citet{ilharco2022patching}, the weights of the classification layer, which are derived from encoding a standard set of zero-shot template prompts for each dataset, are frozen during fine-tuning. Freezing this layer ensures no additional learnable parameters are introduced and does not negatively affect accuracy \citep{ilharco2022patching}.
Regarding the language experiments, we aligned with \citet{yadav2023ties, ilharco2023editing} and utilized three variants of the T5 model \citep{2020t5}, namely T5-Small, T5-Base, and T5-Large, with training conducted for a maximum of 75,000 steps. We employed an effective training batch size of 1024, with a learning rate of $10^{-4}$. To prevent overfitting, we implemented an early stopping mechanism with a patience threshold of 5. During training, we used \texttt{bfloat16} and the maximum sequence length was set to 128. Evaluation is carried out by performing rank classification, where the model's log probabilities for all possible label strings are ranked. The prediction is considered correct if the highest-ranked label corresponds to the correct answer.

\textbf{Disentanglement error heatmaps.} As prescribed by \citet{Tangent_task_arith_2023}, we produce the weight disentanglement visualizations of Figure \ref{fig:weight_disentangle} by computing the value of the disentanglement error $\xi(\alpha_1, \alpha_2)$ on a $20 \times 20$ grid of equispaced values in $[-3, 3] \times [-3, 3]$. Estimations are carried out on a random subset of 2,048 test points for each dataset.

\textbf{Tuning of $\alpha$ in Task Arithmetic experiments.} As outlined in \citet{ilharco2023editing, Tangent_task_arith_2023}, we employ a single coefficient, denoted as $\alpha$, to adjust the size of the task vectors used to modify the pre-trained models (\ie $\alpha_1 = \alpha_2 = ... \alpha_t$). For both the task addition and task negation benchmarks, following fine-tuning, we evaluate different scaling coefficients from the set $\alpha \in \{0.0, 0.05, 0.1, ..., 1.0\}$ and select the value that achieves the highest target metric on a small held-out portion of the training set, as specified in \citet{ilharco2023editing, Tangent_task_arith_2023}. 
To account for the lower norm of task vectors obtained via sparse fine-tuning (LoTA and \shortname) we extend this range by $\times 1 / (1-k)$ where $k$ is the sparsity ratio of the task vector.
Specifically, we aim to maximize the \emph{normalized average accuracy} for Task Addition and ensure the minimum \emph{target accuracy} for Task Negation while maintaining at least 95\% of the original accuracy of the pre-trained model on the control task. The tuning of $\alpha$ is performed independently for each method.

\textbf{Measuring computational costs and memory footprint.} The timings in Table \ref{tab:efficiency} are obtained using the \texttt{perf\_counter} clock from Python’s \texttt{time} module. We monitored memory footprint using the NVIDIA \texttt{nvml} library \footnote{\url{https://docs.nvidia.com/deploy/nvml-api/}}. All measurements are obtained during fine-tuning, with the very same setup explained in the fine-tuning details. Then, for each method, the mean and standard deviation of the timings are computed over all iterations of all tasks. Peak memory usage, instead, is taken as the maximum over all tasks. Memory usage is recorded at regular intervals of 1 second, starting from the first forward pass and ending when the training loop breaks.

\textbf{Normalized accuracy calculation in Task Addition.} \emph{Normalized accuracy} is computed by taking the average of the normalized individual accuracies over the $T$ tasks. Given a task $t$, the normalized individual accuracy for $t$ is computed by taking the accuracy of the multi-task fused model on $t$ and dividing it by the single-task accuracy that the fine-tuned checkpoint obtained on $t$ before being fused. Formally,
\begin{equation}
    \text{Normalized Accuracy} = \frac{1}{T} \sum_{t=1}^{T} \frac{\text{Accuracy}[f(\mathcal{D}_t, \bm{\theta}_0 + \sum_{t'}^{T}\alpha_{t'}\bm{\tau}_{t'})]}{\text{Accuracy}[f(\mathcal{D}_t, \bm{\theta}_0 + \alpha_{t}\bm{\tau}_{t})]}
\end{equation}

\begin{algorithm}[h!]\small
\caption{\shortname to obtain task vectors}\label{alg:ours_sparse_ft}
\SetKwInOut{Input}{Input}
\SetKwInOut{Output}{Output}
\SetKw{Return}{return}

\Input{Pre-trained model $\bm{\theta}_0 \in \mathbb{R}^m$, neural network $f(\bm{x}, \bm{\theta}) \triangleq \log p_{\bm{\theta}}(y|\bm{x})$, task dataset $\mathcal{D}_t$, final sparsity $k$, number of rounds $R$, number of epochs $E$, learning rate $\gamma$, loss function $\mathcal{L}$}
\Output{Task vector $\bm{\tau}_t \in \mathbb{R}^m$ for performing task arithmetic}

\emph{// Calibrate sparse fine-tuning mask} \\
$\bm{c} \gets \mathbbm{1}$ \Comment{Initialize weight mask to all 1s}\\
\For{$r=1,2,...,R$}{
$p \gets k^{(r / R)}$ \Comment{Compute the current sparsity at round $r$}\\
$\bm{s} \gets \bm{0}$ \Comment{Initialize parameter-wise scores to all 0s}\\
\emph{// Compute diagonal FIM score according to Equation \ref{eq:diag_fim}} \\
\For{$\bm{x} \in \mathcal{D}_t$}{

    $y \sim p_{(\bm{c} ~\odot~ \bm{\theta}_0)}(y|\bm{x})$ \Comment{Sample from output distribution of the model}

    $\bm{s} \gets \bm{s} + [\nabla_{\bm{\theta}} \log p_{(\bm{c} ~\odot~ \bm{\theta}_0)}(y|\bm{x})]^2 $ \Comment{Update scores on current example \& sampled $y$}
}

\emph{// Update $\bm{c}$ to retain only the bottom-$k$ parameters} \\
$\hat{\bm{s}} \gets \text{\texttt{sort\_descending($\bm{s}$)}}$ \Comment{Sorted scores in descending order} \\
$p \gets \lfloor p \cdot m \rfloor$ \Comment{compute bottom-$p$ threshold index} \\
\For{$j = 1,2,...,m$}{
    \If{$\bm{s}_{[j]} - \hat{\bm{s}}_{[p]} > 0$}{
        $\bm{c}_{[j]} \gets 0$ \Comment{Set the mask of the $j$-th parameter to zero} \\
    }
}
}

\emph{// Sparse fine-tuning, starting from $\bm{\theta}_0$ and obtaining $\bm{\theta}_t^\star$} \\
\For{$\text{epoch} = 1,2,...,E$}{
    \For{$(\bm{x}, y) \in \mathcal{D}_t$}{ 
        $\bm{\theta} \gets \bm{\theta} - \gamma ~[\bm{c} \odot \nabla_{\bm{\theta}}\mathcal{L}(f(\bm{x},\bm{\theta}),y)]$ \Comment{Update rule, mask gradients with $\bm{c}$} \\
    }
}
$\bm{\tau}_t \gets \bm{\theta}_t^\star - \bm{\theta}_0$ \Comment{Compute final task vector for task $t$} \\
\Return{$\bm{\tau}_t$}
\end{algorithm}

\subsection{Details on Mask Calibration \& Computational Overhead}\label{sec:mask_calibration}

Sparse fine-tuning prescribes to mask gradients when updating the model parameters. Thus, it is foundational that the mask is correctly calibrated before training. We mask only Linear, Attention, LayerNorm, and Convolutional layers \citep{kwon2022fast}. Embedding layers and final projection layers are kept frozen. Furthermore, following standard procedures in Pruning-at-Initialization (PaI) \citep{tanaka2020synflow, wang2023ntksap}, we iteratively refine the mask in multiple rounds to obtain better estimates from the mask calibration procedures. In detail, at each round, we select the bottom-$p$ parameters (according to our score, detailed in Section \ref{sec:method}) and we exponentially increase the current sparsity $p$. We repeat this process until we reach the target sparsity $k$. 
For the sake of major clarity, we report in Algorithm \ref{alg:ours_sparse_ft} the pseudocode for our procedure, encompassing both mask calibration and sparse fine-tuning.
We remark that the choice of the bottom-$k$ values may lead to \emph{layer collapse} \citep{tanaka2020synflow}, namely, removing all parameters in a layer, disrupting the information flow in the network. To face this problem, we set $\bm{c}$ to some positive value close to zero (\eg 0.01) and we don't include in the ranking those entries that are already soft-masked. This ensures that we are not changing the nature of our estimation, while countering the possibility of disrupting gradient flow in the network, during calibration.

Unfortunately, mask calibration introduces some amount of overhead before training. It is of paramount importance that such overhead doesn't hinder the computational gains obtained during fine-tuning.

\begin{table}[h!]
    \centering
    \setlength{\aboverulesep}{0pt}
    \setlength{\belowrulesep}{0pt}
    \setlength{\extrarowheight}{.75ex}
    \resizebox{.98\linewidth}{!}{\begin{tabular}{l | a a | a | b b | b || a a b b}%
    \toprule

    \multicolumn{1}{c|}{Method} & \multicolumn{3}{a|}{Avgerage Execution Time (s)} & \multicolumn{3}{b||}{Peak Memory Usage (GiB)} & \multicolumn{2}{a}{Task Addition} & \multicolumn{2}{b}{Task Negation} \\
    \cline{2-7}
    
    {} & Mask & Train & Total & Mask & Train & Overall & Abs. ($\uparrow$) & Norm. ($\uparrow$) & Targ. ($\downarrow$) & Cont. ($\uparrow$) \\
    \midrule

    Non-linear FT \citep{ilharco2023editing} & - & 2479.99 & 2479.99 & - & 18.6 & 18.6 & 86.09 & 90.14 & 20.61 & 72.72 \\
    Linearized FT \citep{Tangent_task_arith_2023} & - & 3311.77 & 3311.77 & - & 21.3 & 21.3 & \ul{88.29} & \ul{93.01} & \ul{10.86} & 72.43 \\
    %\hline\hline
    L-LoRA \citep{tang2023parameter} & - & \ul{1053.07} & \ul{1053.07} & - & \ul{9.7} & \ul{9.7} & 87.77 & 91.87 & 19.39 & 73.14 \\
    LoTA \citep{panda2024lottery} & \tb{51.84} & 2592.40 & 2644.24 & 12.9 & 15.4 & 15.4 & 87.60 & 91.89 & 22.02 & \ul{73.22} \\
    %\midrule
    \textbf{\shortname (Ours)} & 63.04 & \tb{656.23} & \tb{719.27} & \tb{7.8} & \tb{7.8} & \tb{7.8} & \tb{88.40} & \tb{95.19} & \tb{10.63} & \tb{73.55} \\

    \bottomrule
    \end{tabular}%
    }
    %\vspace{1.5mm}
    \caption{\textbf{Computational cost and memory footprint of mask calibration and fine-tuning.} Average time (in seconds) and peak memory usage (in Gibibytes) of mask calibration and fine-tuning approaches on CLIP ViT-L/14, alongside their performance on the task arithmetic benchmark. For both LoTA and \shortname, we used batch size 128 for 40 iterations (in detail, 10 iterations per round for \shortname, with 4 rounds total). We employ gradient checkpointing during mask calibration. Further details on the resource monitoring process can be found in Appendix \ref{sec:implementation}.
    \textbf{Bold} indicates the best results. \underline{Underline} the second best.}
    \label{tab:rebuttal_efficiency}
    \vspace{-3mm}
\end{table}

\begin{figure}[h!]
    \centering
    \includegraphics[width=\linewidth]{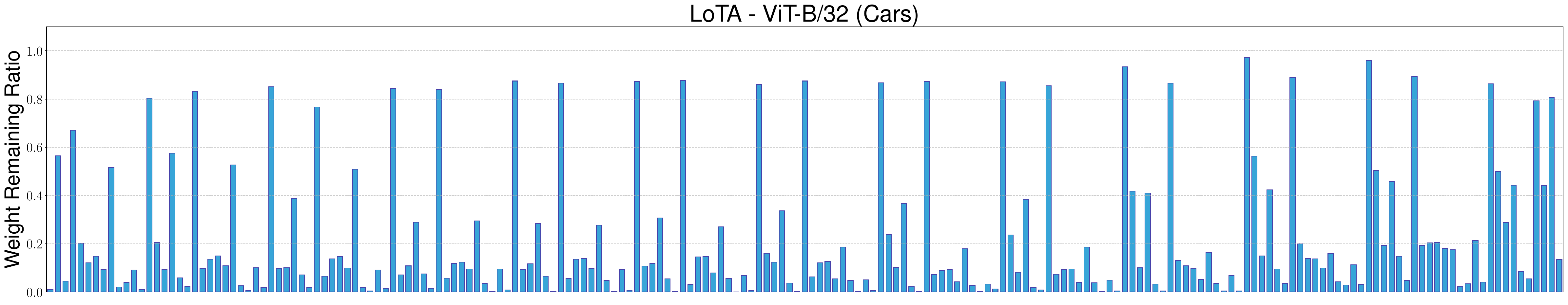}
    \vspace{12pt}
    \includegraphics[width=\linewidth]{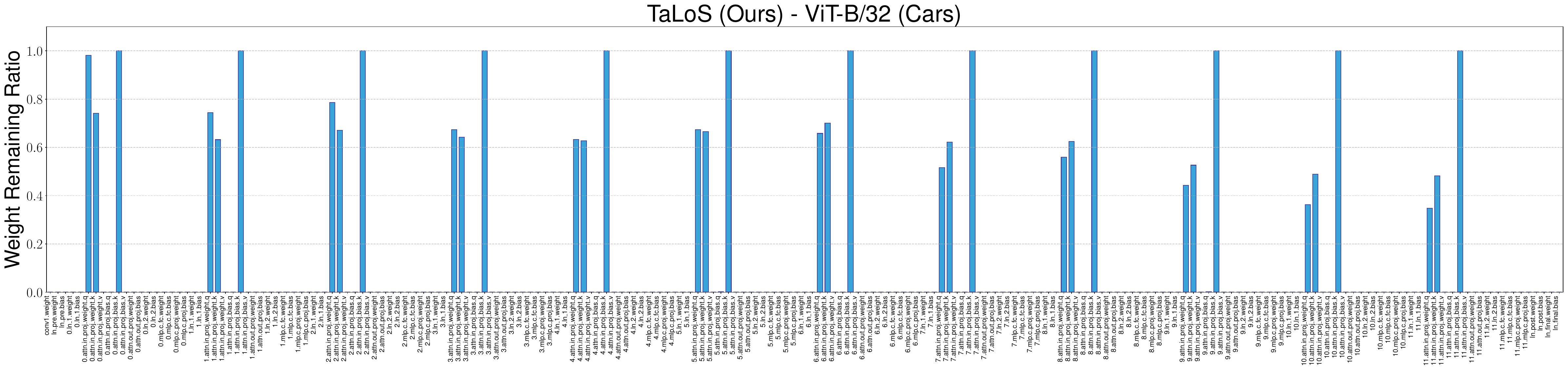}

    \includegraphics[width=\linewidth]{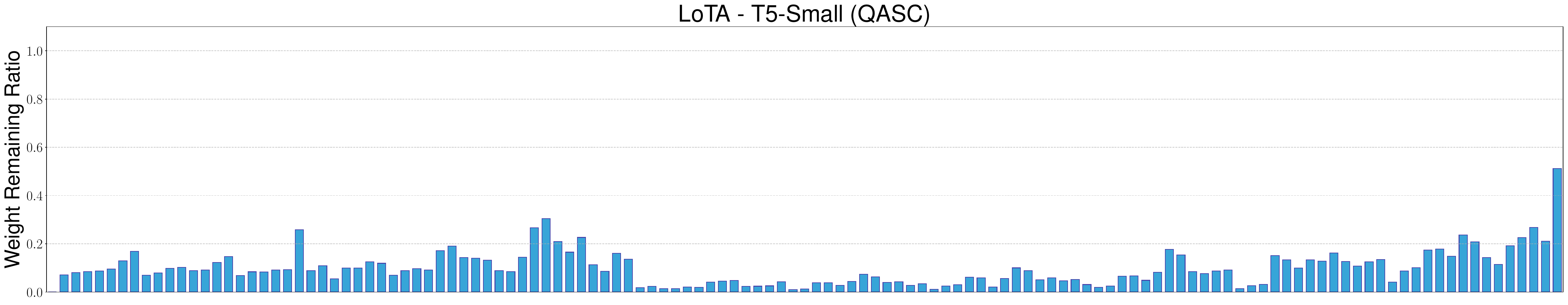}
    \includegraphics[width=\linewidth]{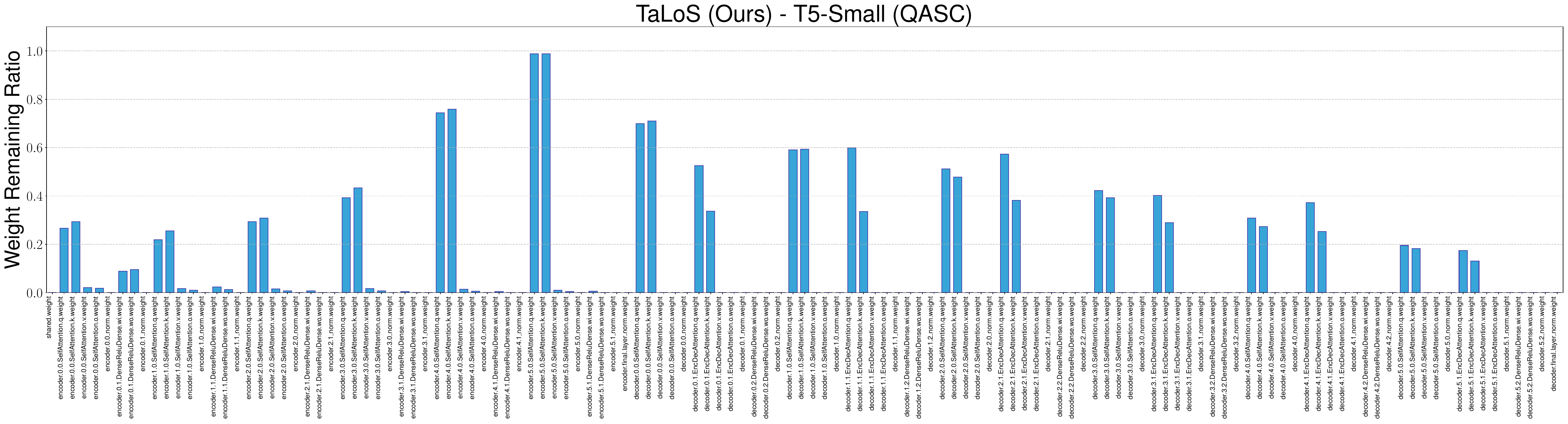}
    
    \vspace{-2.5mm}\caption{\textbf{Visualization of mask calibration.} Percentage of parameters selected for sparse fine-tuning in a ViT-B/32 (top) and a T5-Small (bottom) models, after our method's mask calibration vs. LoTA's mask calibration, at 90\% sparsity. On ViT-B/32, we calibrate the masks on the Cars dataset \citep{krause2013cars}, while on T5-Small we use QASC \citep{khot2020qasc}.}
    \label{fig:sparsity}
\end{figure}

\textbf{Time overhead.} The time spent for a single iteration of mask calibration is comparable to that of a single forward-backward iteration of non-linear fine-tuning (refer to Table \ref{tab:efficiency}). Our mask calibration process typically employs an average of 10 iterations per round, with satisfactory results already observed at just 4 rounds (\ie, approximately 40 iterations total, we use the same batch size for mask calibration as for fine-tuning). Given that fine-tuning generally requires around 2,000 iterations for vision experiments and substantially more for language tasks, we argue that the time overhead introduced by our mask calibration is negligible. 

\textbf{Memory overhead.} The memory cost of each mask calibration iteration is equivalent to that of each training iteration in non-linear fine-tuning. While we have not implemented any specific mechanism to reduce the memory footprint for calculating gradients (used as scores) during mask calibration, there are several approaches available to achieve this. Most of these methods involve estimating gradients using zeroth-order information \citep{hinton2022forward, malladi2023fine, sung2023ecoflap}, which allows to trade off speed for reduced memory usage by approximating gradients through multiple forward passes, eliminating the need to store computational graphs for automatic differentiation. Alternatively, gradient checkpointing \citep{chen2016training} is another practical solution.

To further clarify the overall computational cost of \shortname, encompassing both mask calibration and sparse fine-tuning, we provide a comparison in Table \ref{tab:rebuttal_efficiency} of the timings in seconds (averaged over the 8 vision tasks) and the peak memory usage in Gibibytes of mask calibration and fine-tuning on a CLIP ViT-L/14. The results show that mask calibration time is approximately the same for \shortname and LoTA, however, the costs in terms of memory are very different (LoTA requires storing optimizer states). Regarding total time, we recover what was presented in Table \ref{tab:efficiency}, highlighting the beneficial effect of the highly structured sparsity of \shortname on fine-tuning. The task arithmetic results are in line with Tables \ref{tab:task_addition}, \ref{tab:task_negation}, with no detrimental effect given by the usage of gradient checkpointing.

\subsection{Full Mask Calibration Visualizations} \label{sec:full_mask_vis}

For the sake of completeness, we provide a full visualization in Figure \ref{fig:sparsity} of the masks obtained after calibration with \shortname and LoTA. As shown, a repeating sparsity pattern emerges for our method across each transformer block. Notably, \shortname consistently identifies only the \tb{Q} and \tb{K} parameters for fine-tuning, demonstrating a more structured behavior. In contrast, the mask generated by LoTA appears far more unstructured, with no clear pattern across the blocks.

\subsection{Analyzing the Fine-tuning Behavior}\label{sec:fixed_features}

\begin{figure}[h!]
    \centering
    \includegraphics[width=0.98\linewidth]{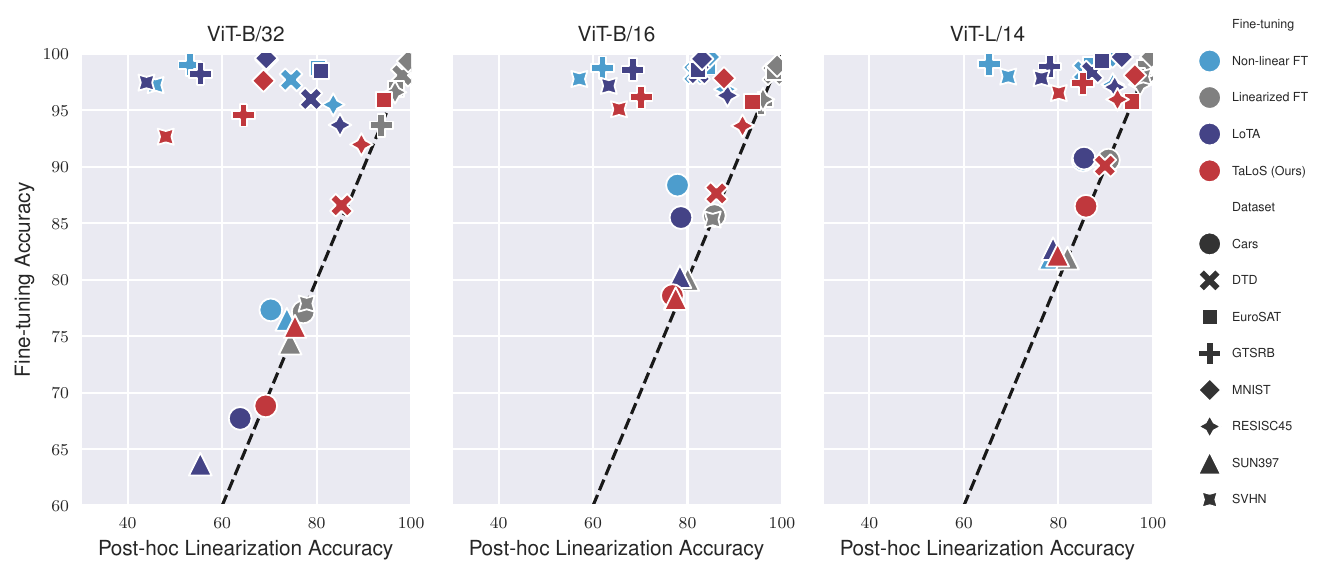}
    
    \vspace{-2.5mm}\caption{\textbf{Testing linearized behavior.} Single-task accuracies of different fine-tuning strategies, each used to obtain their corresponding task vectors $\bm{\tau}_t$, and the accuracy of their post-hoc linearization $f_\text{lin}(\cdot, \bm{\theta}_0 + \bm{\tau}_t)$. Different colors represent distinct fine-tuning strategies, while different markers indicate different tasks. Points that lie on the bisector (black dashed line) indicate that the fine-tuning process exhibited linearized behavior.}
    \label{fig:linear_advantage}
\end{figure}
\begin{figure}[h!]
    \centering
    \includegraphics[width=\linewidth]{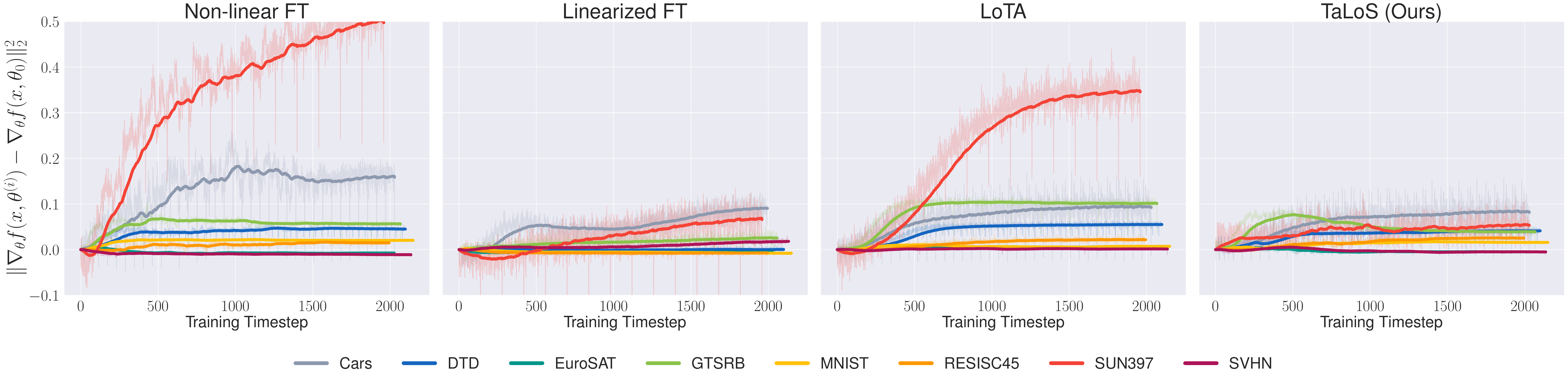}
    
    \vspace{-2.5mm}\caption{\textbf{Change in parameter sensitivity throughout fine-tuning.} We visualize the average relative change in the output derivative of the parameters of a CLIP ViT-B/32 model when fine-tuned using different approaches. The starting point is the same for all methods.}
    \label{fig:fixed_features}
    \vspace{-3mm}
\end{figure}

We provide an empirical validation on the linear fine-tuning regime of our \shortname (\ie the change in the network output can be well-approximated by its first-order Taylor expansion around $\bm{\theta}_0$). 
As discussed by \cite{Tangent_task_arith_2023}, a cheap test consists of performing post-hoc linearization of the fine-tuned model around $\bm{\theta}_0$ and checking whether the performance produced by such a linearized model matches that of the original fine-tuned model. We use this approach and report the results in Figure \ref{fig:linear_advantage}. 
The scatter plots compare the fine-tuning accuracy against the post-hoc linearization accuracy for various tasks and fine-tuning strategies across different ViT architectures. Our method, \shortname, consistently demonstrates linearized behavior during fine-tuning for most tasks, as evidenced by its proximity to the bisector line. This supports our claim that sparse fine-tuning, which both \shortname and LoTA employ, inherently promotes the emergence of linearized behavior during fine-tuning. Interestingly, while \shortname exhibits this property across a wide range of tasks, LoTA does not consistently demonstrate the same level of linearized behavior. This discrepancy can be attributed to differences in parameter selection, as discussed in the next paragraph, closely matching what happens during linearized fine-tuning. It's worth noting that linearized behavior may arise for various fine-tuning strategies, but its occurrence depends on the interaction between the task and pre-training \citep{malladi2023kernel}. For instance, tasks such as GTSRB \citep{stallkamp2011gtsrb}, MNIST \citep{lecun1998mnist}, and SVHN \citep{netzer2011svhn} do not exhibit fine-tuning in the linear regime, hinting at a potential mismatch with the pre-training, as evidence suggests \citep{radford2021clip}.

To further test the fine-tuning regime, we examine the evolution of parameter sensitivity during fine-tuning across different methods, as depicted in Figure \ref{fig:fixed_features}. 
Inspired by \citet{malladi2023kernel}, we measure the average change in sensitivity as $\mathbb{E}_{\bm{x}} [\| \nabla_{\bm{\theta}} f(\bm{x}, \bm{\theta}^{(i)}) - \nabla_{\bm{\theta}} f(\bm{x}, \bm{\theta}_0)\|_2^2]$ at each $i$-th training step, with $\bm{x}$ from a small subset of 2,048 examples from $\mathcal{D}_t$.
Notably, for \shortname, the gradient $\nabla_{\bm{\theta}} f(\bm{x}, \bm{\theta})$ remains almost unchanged throughout training, closely mirroring the behavior of linearized fine-tuning. In contrast, LoTA diverges from this pattern, behaving more in line with non-linear fine-tuning. This phenomenon reinforces our claim that our method fine-tunes in the linearized regime, as maintaining a constant $\nabla_{\bm{\theta}} f(\bm{x}, \bm{\theta})$ during fine-tuning is critical for operating in the linearized regime \citep{malladi2023kernel}.

%%%%%%%%%%%%%%%%%%%%%%%%%%%%%%%%%%%%%%%%%%%%%%%%%%%%%%%%%%%%%%%%

\subsection{Ablations on Mask Sparsity Ratio}\label{sec:k_ablation}

\begin{figure*}[h!]
    \centering
    \includegraphics[width=0.245\linewidth]{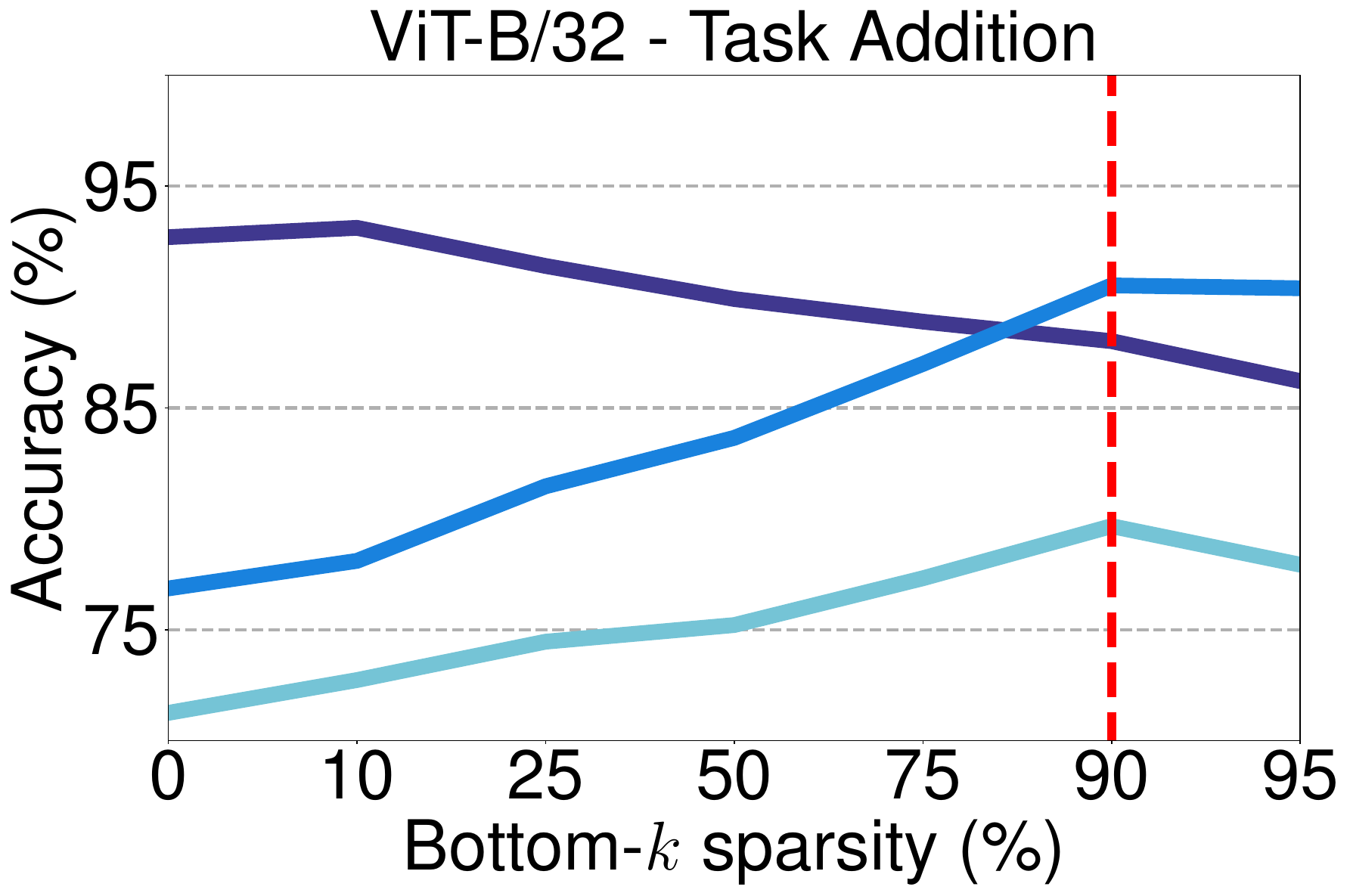}
    \includegraphics[width=0.245\linewidth]{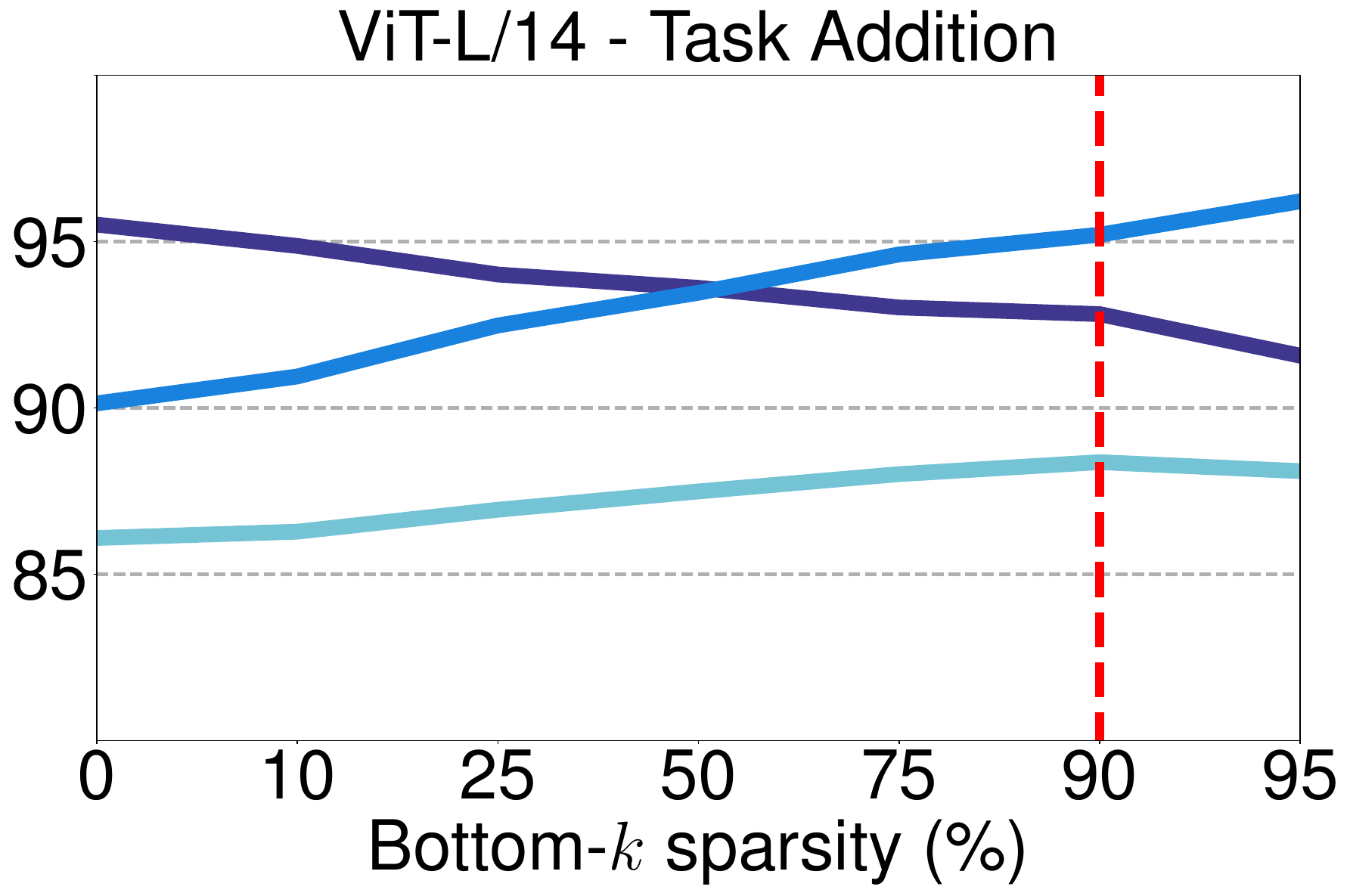}
    \includegraphics[width=0.245\linewidth]{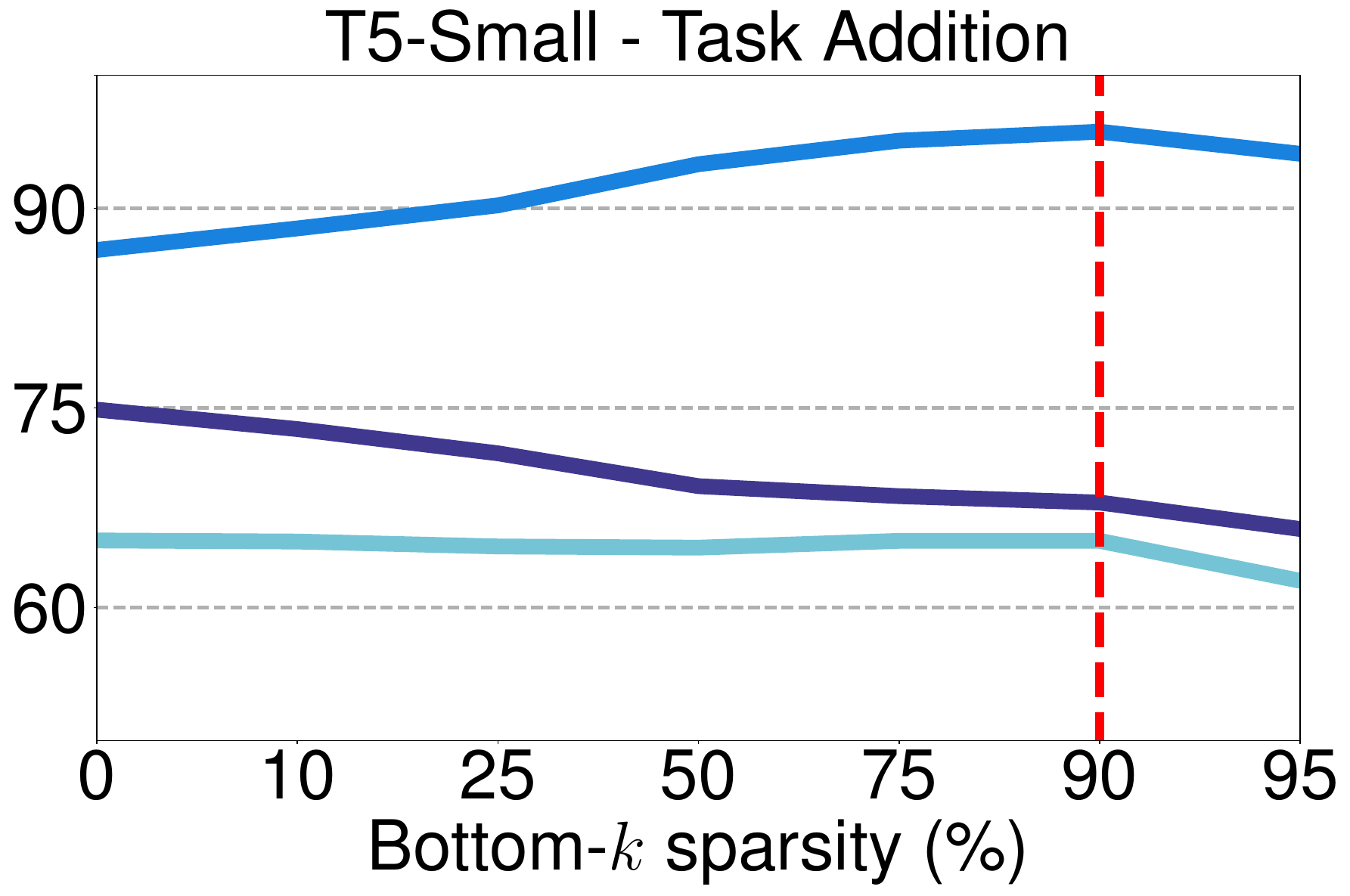}
    \includegraphics[width=0.245\linewidth]{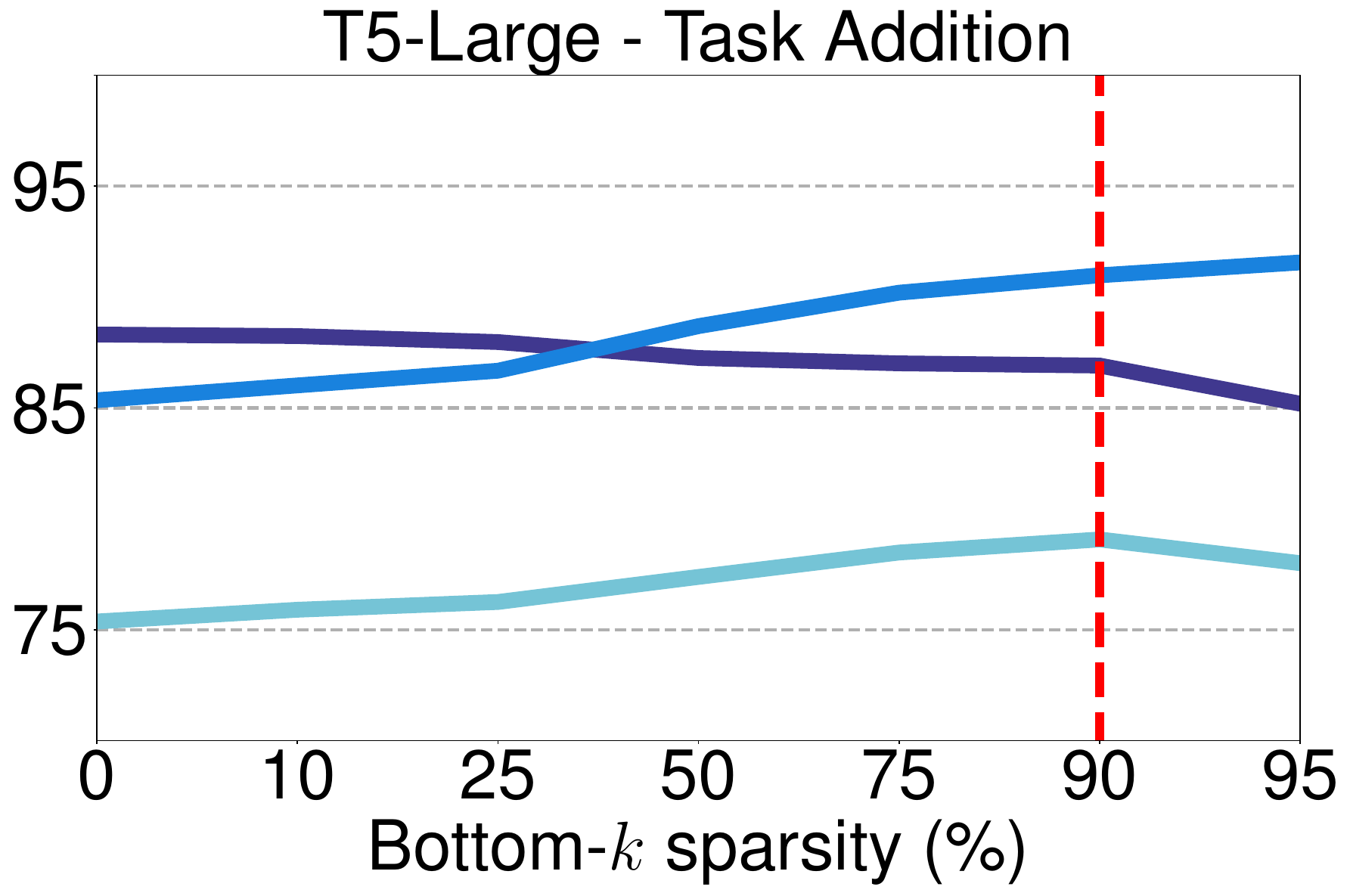}

    \includegraphics[width=0.6\linewidth]{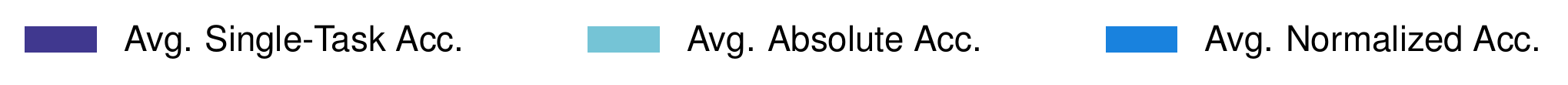}
    \vspace{5mm}

    \includegraphics[width=0.245\linewidth]{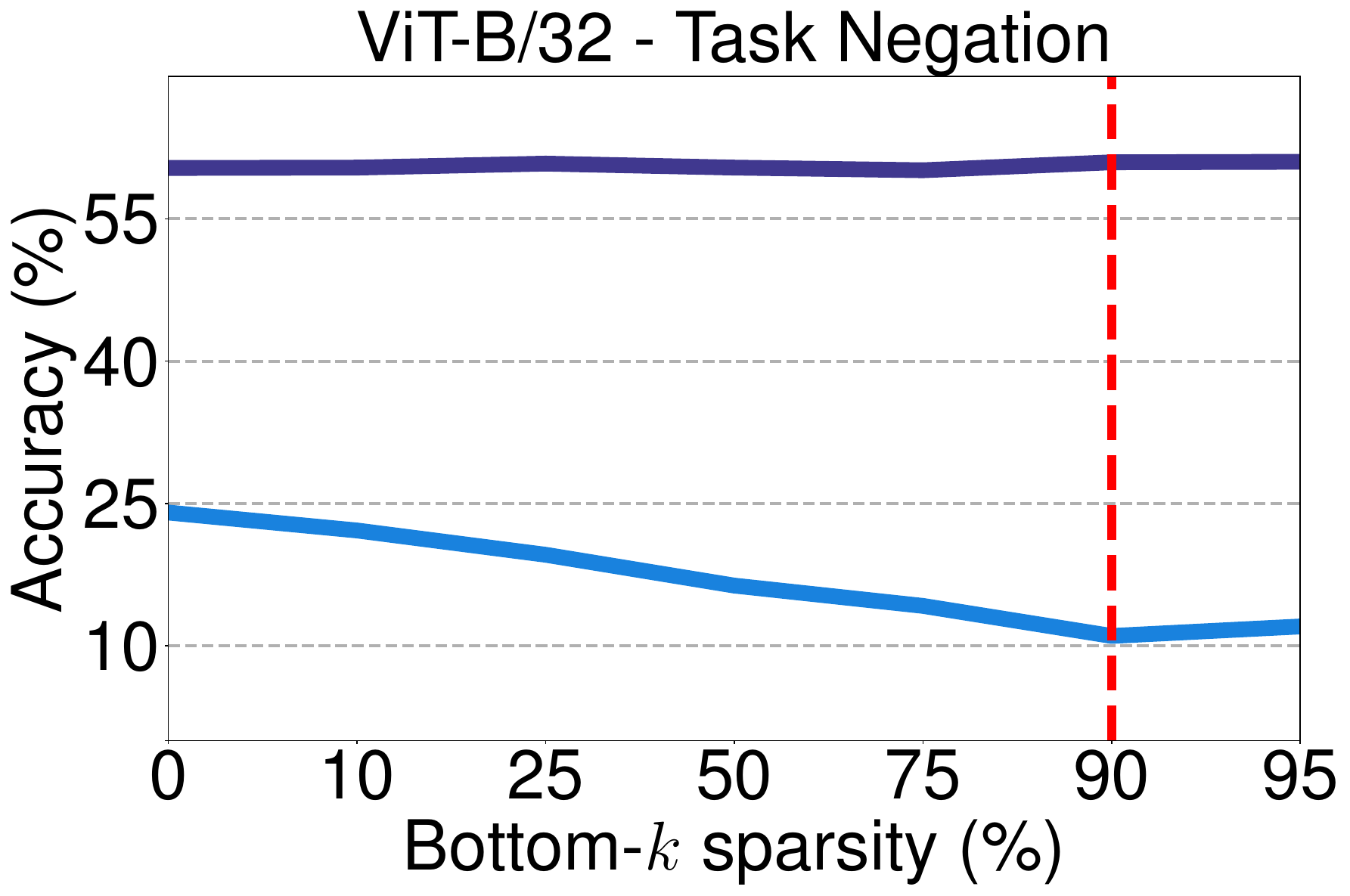}
    \includegraphics[width=0.245\linewidth]{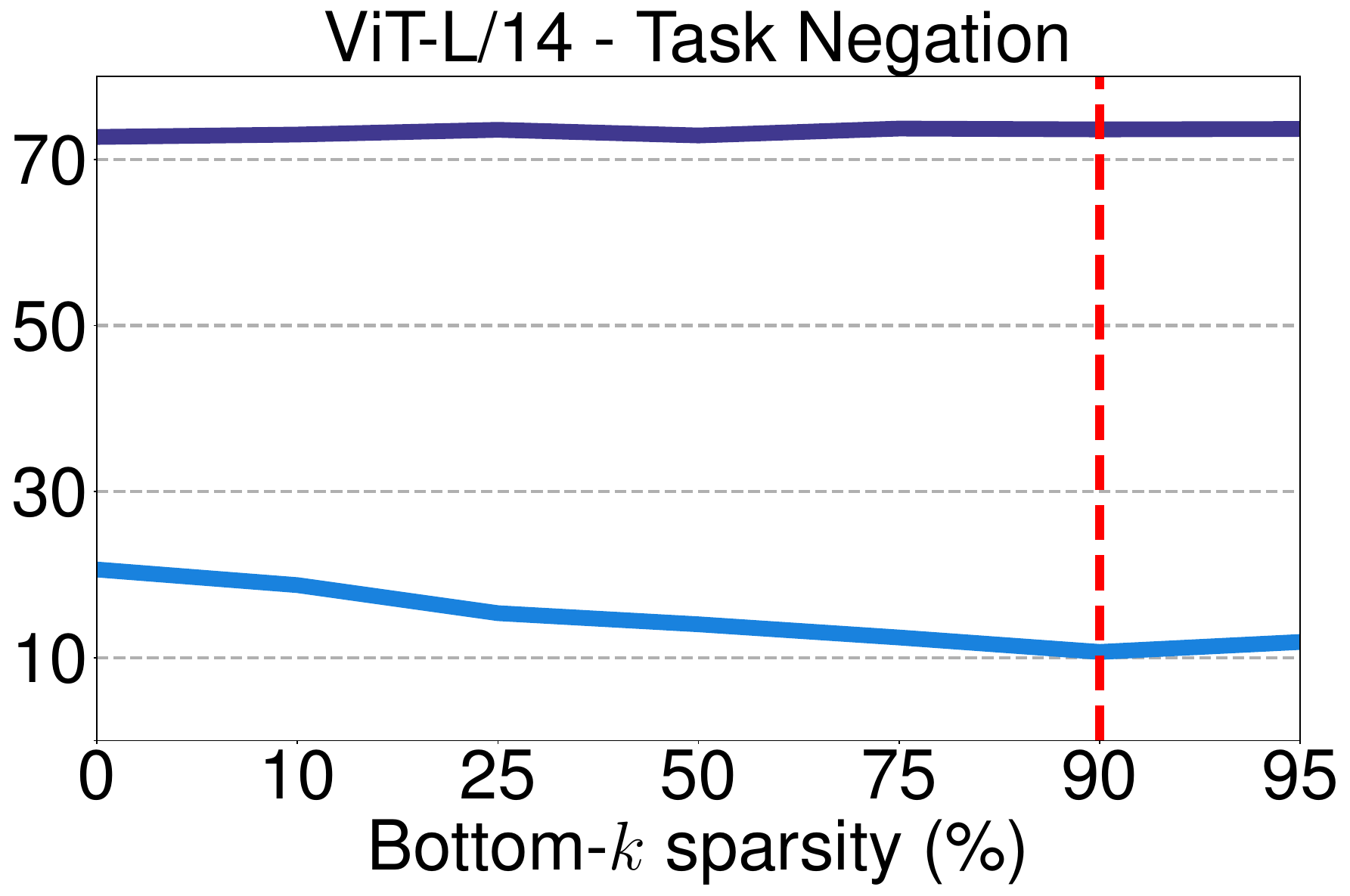}
    \includegraphics[width=0.245\linewidth]{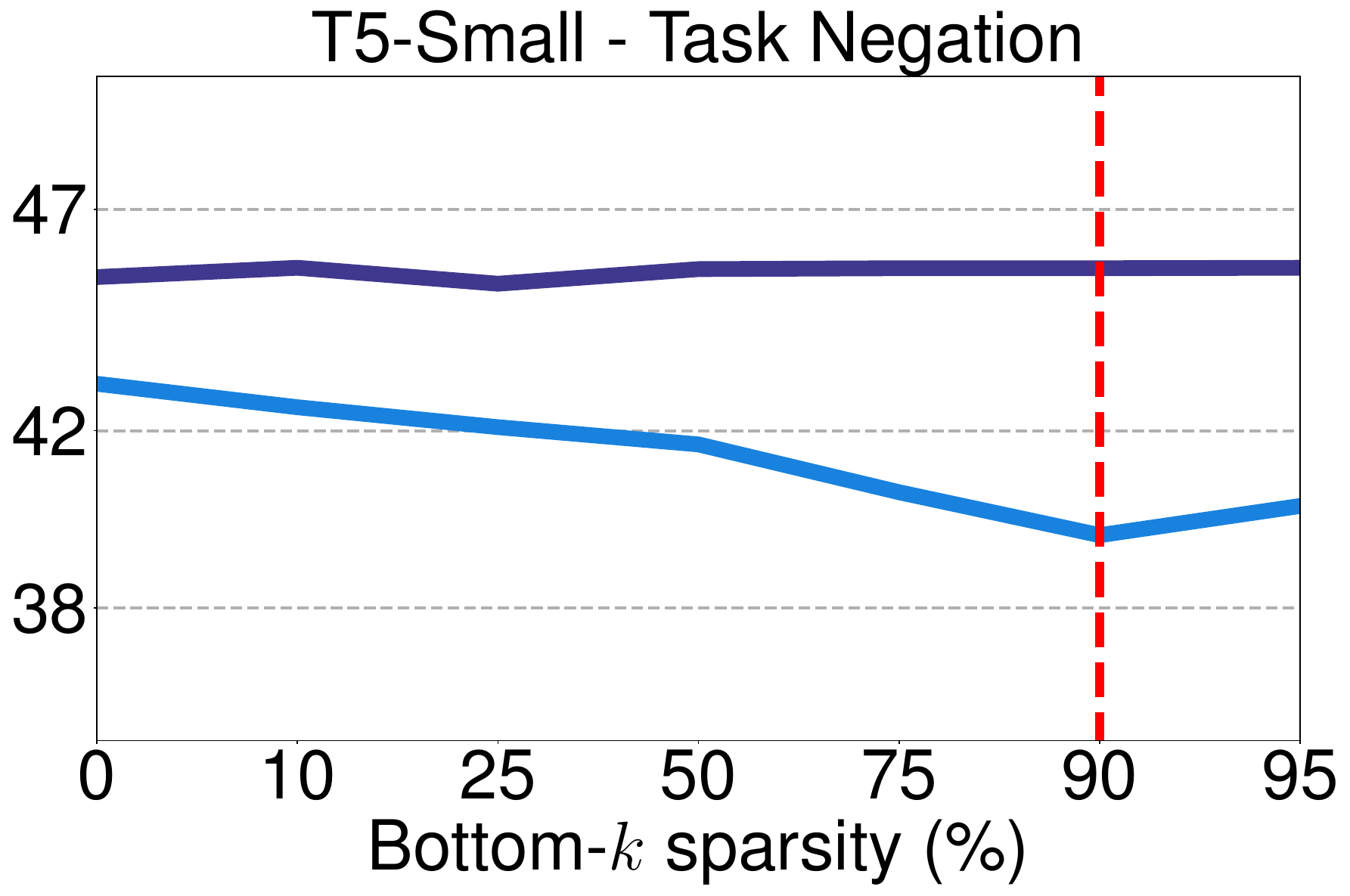}
    \includegraphics[width=0.245\linewidth]{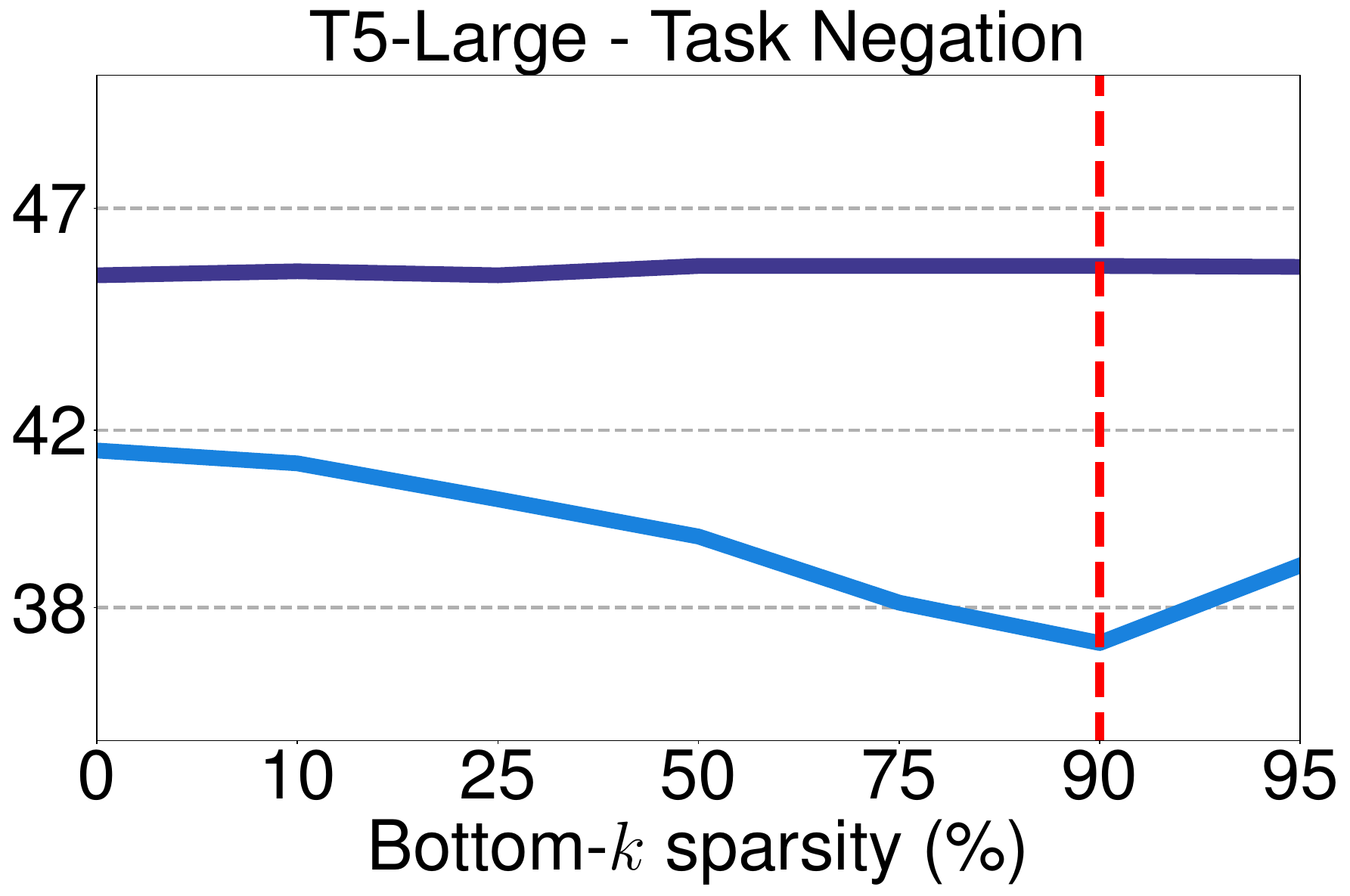}
    
    \includegraphics[width=0.425\linewidth]{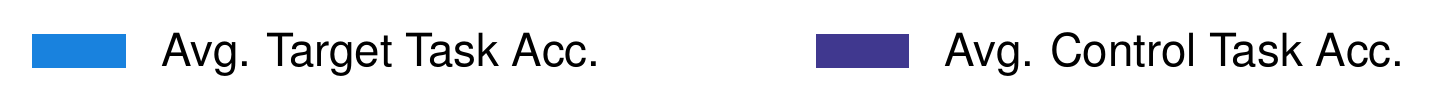}
    
    \vspace{-2.5mm}\caption{\textbf{Effect of the choice of $k$ in \shortname.} Results of hyperparameter tuning of $k$ in \shortname for task addition and negation on both vision and language. Note that we tune $k$ indirectly by controlling its value via the sparsity ratio. For \tb{task addition} (top) we report the average single-task accuracy (before addition), absolute and normalized accuracies (after addition). For \tb{task negation} (bottom) we report average target and control accuracies (after negation).
    }
    \label{fig:calib_bottom_k}
    \vspace{-3mm}
\end{figure*}

\begin{figure*}[t!]
    \centering
    \includegraphics[width=0.82\linewidth]{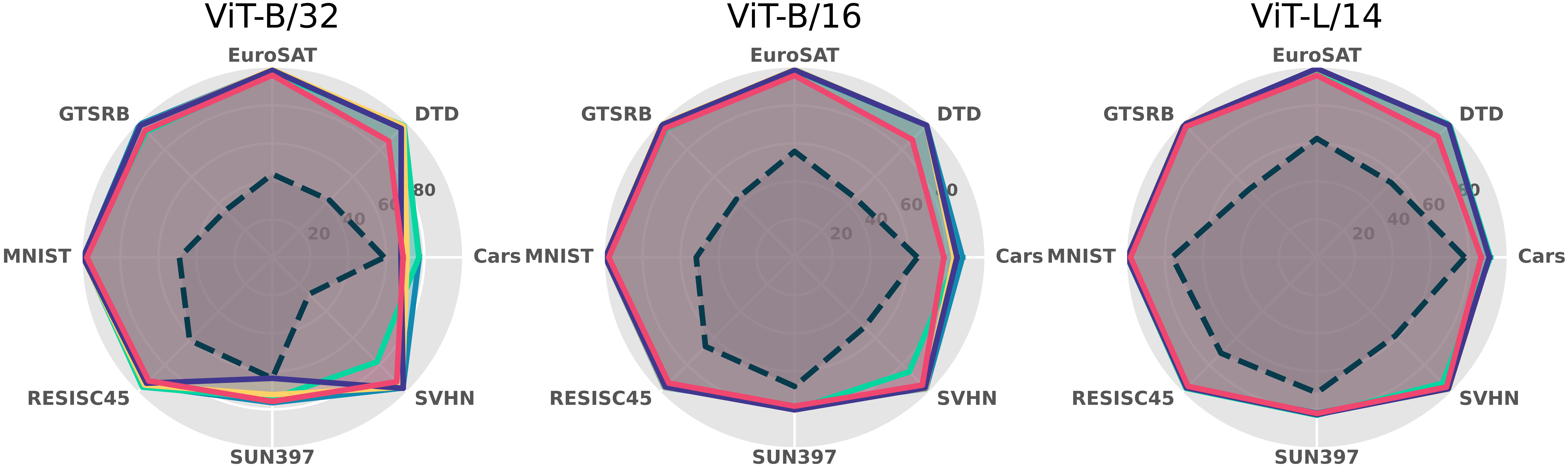}
    
    \vspace{12pt}
    
    \includegraphics[width=\linewidth]{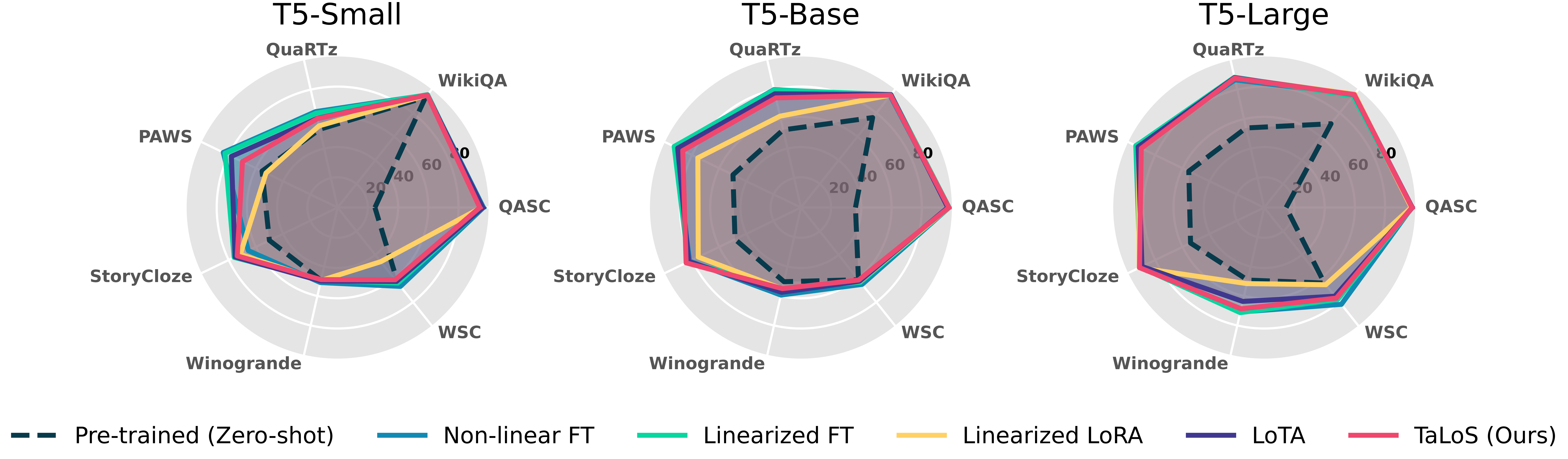}

    \vspace{-2.5mm}\caption{\textbf{Task performance after fine-tuning.} Single-task accuracies obtained by different fine-tuning approaches across vision and language experiments. Results are displayed for three model sizes of CLIP ViT (B/32, B/16, L/14) and T5 (Small, Base, Large), with outer edges representing higher accuracy. The dashed line represents the accuracies before fine-tuning.
    }
    \label{fig:polar_singletask}
    \vspace{-3mm}
\end{figure*}

For a clear understanding of the effect of sparsity on \shortname, we report in Figure \ref{fig:calib_bottom_k} the task arithmetic performance achieved by \shortname, while varying the sparsity level. At 0\% sparsity, we recover full (non-linear) fine-tuning results. Increasing the sparsity improves the task arithmetic performance, while slightly decreasing the average single-task accuracy, as fewer parameters are updated during fine-tuning. Optimal values for absolute accuracy (in task addition) and target accuracy (in task negation) are observed for a sparsity level of 90\% across a variety of models. After 90\% sparsity there is a slight drop in both task arithmetic and single-task performance, making such sparsity levels not ideal. Intuitively, if the fine-tuning involves too little weight the resulting entries in the task vector will be mostly zero, reducing the ability to perform task arithmetic effectively.
We can conclude that, like other parameter-efficient fine-tuning methods, our approach trades some single-task performance for parameter efficiency. But this trade-off allows also for superior task arithmetic capabilities for \shortname (Tables \ref{tab:task_addition}, \ref{tab:task_negation}) while maintaining competitive single-task accuracy, especially for larger models where the performance drop becomes negligible (Figure \ref{fig:polar_singletask}).

\subsection{Single-task Performance of Fine-tuning Methods}\label{sec:single_task_perf_peft}

In this analysis we focus on discussing the single-task performance of \shortname before task addition.
To this goal, we compare in Figure \ref{fig:polar_singletask} the accuracies obtained by \shortname (at 90\% sparsity) vs. the other fine-tuning strategies.
In almost all cases \shortname achieves approximately the same performance of full fine-tuning methods (Non-linear FT and Linearized FT), occasionally improving over Linearized FT (ViT-B/32 on SVHN), which is remarkable, as \shortname updates only a very small subset of parameters, while full fine-tuning (both linearized and non-linear) updates the whole set of model parameters. Furthermore, compared with parameter-efficient fine-tuning methods, which allows for a truly fair comparison (the parameter count is the same across methods), almost always \shortname improves with respect to Linearized LoRA and matches the performance of LoTA. However, we remark that the task arithmetic performance of \shortname is much higher than the latter (see Tables \ref{tab:task_addition}, \ref{tab:task_negation}).

\subsection{Additional Evidence on the Parameter-sharing Phenomenon}\label{sec:sharing_analysis}

In this section, we provide additional validation of the phenomenon observed in our motivating example, namely that insensitive parameters are consistently shared across tasks. First, we revisit the relationship between parameter sensitivity and the Fisher Information matrix (FIM) \cite{fisher1922mathematical}, highlighting why the FIM serves as a suitable tool for conducting sensitivity analysis. Next, we present further experimental evidence to support the findings of Section \ref{sec:talos}. Specifically, instead of pruning the least sensitive parameters, we analyze the effect of perturbing them and subsequently examine whether masks calibrated on different tasks exhibit significant similarity.

\textbf{Parameter sensitivity analysis and connection to Fisher Information.}
Applying a perturbation $\bm{\theta}_0’ \gets \bm{\theta}_0 + \delta\bm{\theta}_0$ to a subset of the pre-trained weights $\bm{\theta}_0$ and observing no change in the output $f(\bm{x}, \bm{\theta}_0’) \approx f(\bm{x},\bm{\theta}_0)$ intuitively means that those weights have low sensitivity to the task. So, pruning or randomizing them would not affect input-output behavior.
However, there may be a problem in assessing sensitivity via extreme randomizations/perturbations: if “extreme” randomization refers to very high magnitude perturbations (perhaps, additive), then such perturbations will not be suitable to assess the sensitivity of the parameters, as this could potentially move the current solution (parametrized by $\bm{\theta}_0 \in \mathbb{R}^m$) away from the current local optimum, to a distinct region of the loss landscape. Indeed, sensitivity analysis generally refers to “robustness to small perturbation”. This concept, alongside how to perform proper sensitivity analysis on the parameters of a neural network, has been formalized by a rich literature dedicated to applications of information geometry \citep{amari1996neural, chaudhry2018riemannian, pascanu2013revisiting}. Specifically, as shown by \citet{chaudhry2018riemannian, pascanu2013revisiting}, to assess the influence of each weight on the output of a network, we can use the Kullback-Leibler (KL) divergence between the output distribution induced by the original network ($p_{\bm{\theta}_0}$) vs. the one induced by the perturbed network ($p_{\bm{\theta}_0 + \delta\bm{\theta}}$). Mathematically, assuming $\delta\bm{\theta} \rightarrow 0$ (a small perturbation),

\begin{equation*}
    D_\text{KL}(p_{\bm{\theta}_0} \| p_{\bm{\theta}_0 + \delta\bm{\theta}}) = \frac{1}{2} \delta\bm{\theta}^\top F(\bm{\theta}_0) \delta\bm{\theta} + \mathcal{O}(\|\delta\bm{\theta}\|^3)~.
\end{equation*}

The KL divergence is zero if the perturbation doesn’t affect the output, revealing that the modified weights are not influential for the output. It is larger than zero otherwise. Here $F(\bm{\theta}_0) \in \mathbb{R}^{m \times m}$ is the Fisher Information matrix (FIM) \citep{fisher1922mathematical, amari1996neural}. It is a positive semi-definite symmetric matrix defined as,

\begin{equation*}
    F(\bm{\theta}_0) = \mathbb{E}_{\bm{x}} [ \mathbb{E}_{y \sim p_{\bm{\theta}_0}(y|\bm{x})} [\nabla_{\bm{\theta}} \log p_{\bm{\theta}_0}(y|\bm{x}) \nabla_{\bm{\theta}} \log p_{\bm{\theta}_0}(y|\bm{x})^\top ] ]~.
\end{equation*}

It can be used to relate the changes in the parameters to the changes in the outputs, effectively implementing a proper sensitivity analysis of the parameters of a neural network by studying the magnitude of its diagonal elements, as they represent the sensitivity of each parameter \citep{chaudhry2018riemannian, pascanu2013revisiting, mergingfisher_2024}. Formally, for each parameter $j \in 1,...,m$, 
its corresponding entry on the diagonal of the FIM has value

\begin{equation*}
    F_{[j,j]}(\bm{\theta}_0) =\mathbb{E}_{\bm{x}} [ \mathbb{E}_{y \sim p_{\bm{\theta}_0}(y|\bm{x})} [\nabla_{\bm{\theta}_{[j]}} \log p_{\bm{\theta}_0}(y|\bm{x})]^2 ]~.
\end{equation*}

\begin{figure}[t!]
    \centering
    \includegraphics[width=0.6\linewidth]{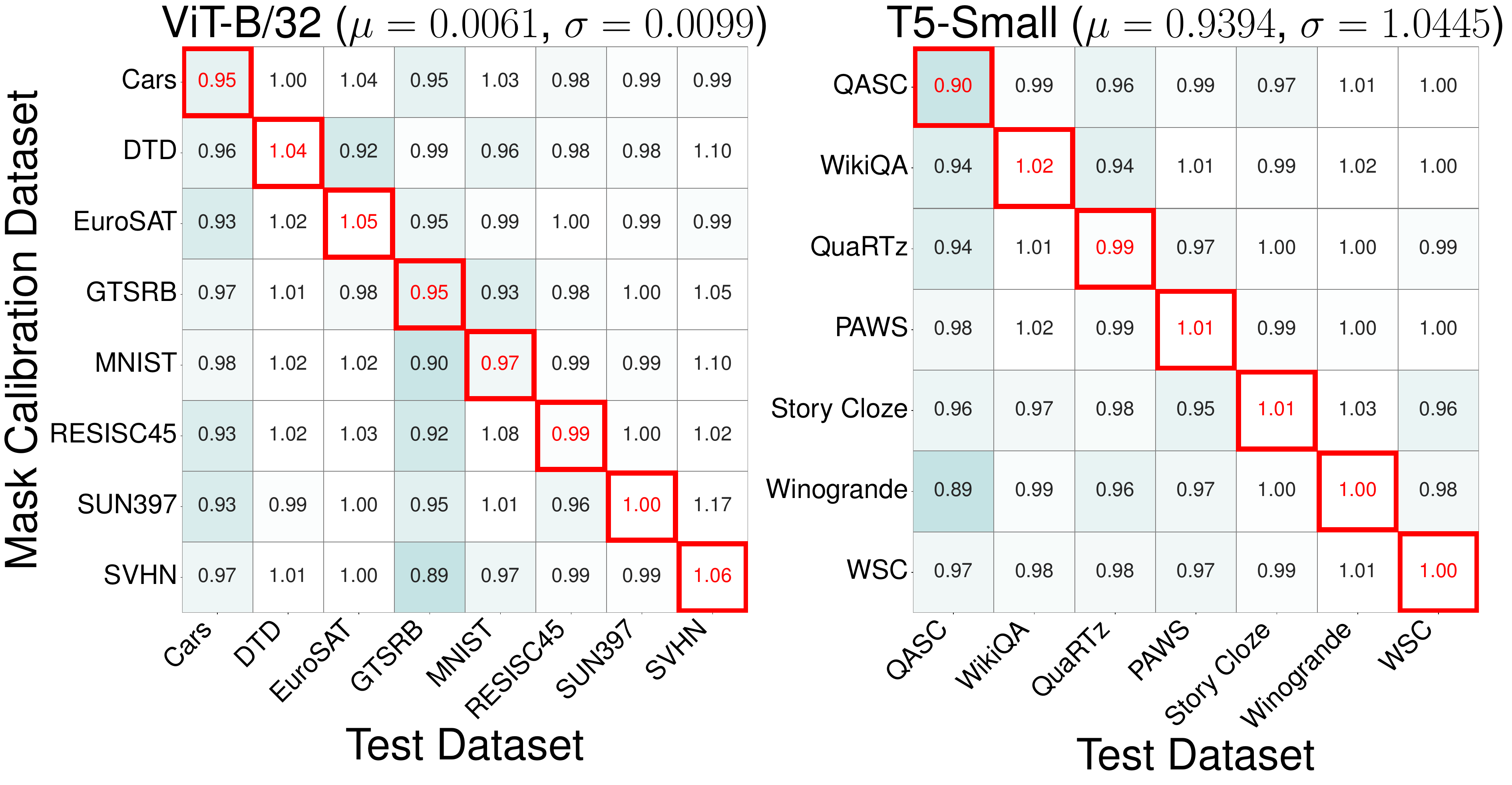}\\
    
    \vspace{-2.5mm}\caption{\textbf{Perturbing parameters with low sensitivity.} The heatmaps illustrate the effect of perturbing the parameters with the lowest sensitivity (measured by $[F_{[j,j]}(\bm{\theta}_0, \mathcal{D}_t)]_{j=1}^m$) on different tasks across various pre-trained models. Each grid compares the accuracy ratios for models after pruning, with the rows representing the task $\mathcal{D}_t$ used to identify the parameters with the lowest sensitivity and the columns showing the model's performance on each task after pruning those parameters. The accuracy ratios are normalized by the model's performance before perturbation. The average magnitude $\mu$ and standard deviation $\sigma$ across perturbed parameters, prior to applying noise are also reported. The ratio of perturbed parameters (10\%) is chosen based on the experiment of Figure \ref{fig:shared_bottomk}.
    }
    \label{fig:shared_bottomk_noise}
    \vspace{-5mm}
\end{figure}

The higher this value, the more the model will be affected by the $j$-th parameter changes.

\textbf{Perturbing the least sensitive parameters.}
We repeat in Figure \ref{fig:shared_bottomk_noise} the experiment of Figure \ref{fig:shared_bottomk}, but by adding noise distributed as $\mathcal{N}(0, 2\sigma I)$ to the bottom-10\% of parameters, instead of pruning them. $\sigma$ is the standard deviation of the parameters, previous to perturbation.
The results align with the analysis reported in Figure \ref{fig:shared_bottomk}, highlighting the stability of these parameters across tasks.

\begin{figure}[t!]
    \centering
    \includegraphics[width=\linewidth]{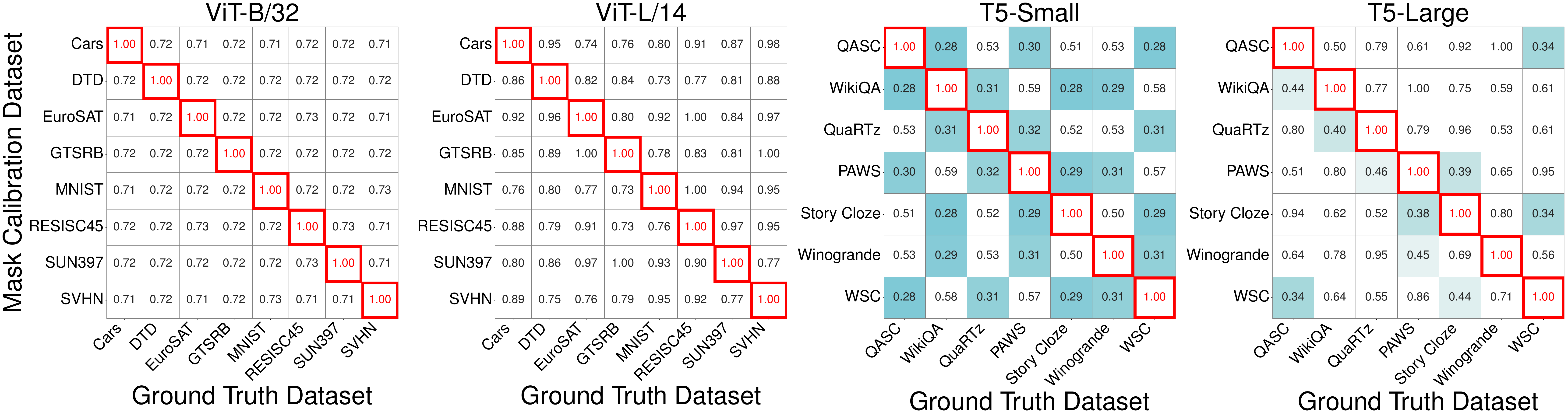}\\
    
    \vspace{-2.5mm}\caption{\textbf{Masks intersections of low sensitivity parameters.} The heatmaps illustrate the mean Intersection over Union (mIoU) between masks pairs of the lowest sensitivity parameters (measured by $[F_{[j,j]}(\bm{\theta}_0, \mathcal{D}_t)]_{j=1}^m$) on all tasks across different pre-trained models. For each mask, the amount of selected parameters (10\%) is chosen based on the experiment of Figure \ref{fig:shared_bottomk}.
    }
    \label{fig:shared_bottomk_miou}
    \vspace{-3mm}
\end{figure}

\textbf{Measuring masks intersections across tasks.}
Additionally, in Figure \ref{fig:shared_bottomk_miou} we provide further evidence about the overlap of low-sensitivity parameters across tasks. For each parameter, we compute the mean Intersection over Union (mIoU) of masks, between each task pair: starting from pre-trained parameters $\theta_0$, we predict the mask on task $t$ and then check its intersection over union against the mask predicted on task $t’$ (which acts as a ground truth). A mIoU of 1 signals perfect mask overlap between tasks. The number of parameters selected by each mask is 10\%, in line with the experiment of Figure \ref{fig:shared_bottomk}. Smaller vision models (ViT-B/32) exhibit high parameter sharing ($>0.7$ mIoU) of low-sensitivity parameters, while smaller language models (T5-Small) share fewer (0.3–0.5 mIoU). However, with a fixed 10\% mask sparsity, larger models in both vision and language domains share more low-sensitivity parameters across tasks.

\begin{table}[t!]
    \centering
    \setlength{\aboverulesep}{0pt}
    \setlength{\belowrulesep}{0pt}
    \setlength{\extrarowheight}{.75ex}
    \resizebox{.68\linewidth}{!}{\begin{tabular}{l a a | b b}%
    \toprule

    \multicolumn{1}{c}{Method} & \multicolumn{2}{a|}{ViT-B/32} & \multicolumn{2}{b}{T5-Small} \\
    {} & Abs. ($\uparrow$) & Norm. ($\uparrow$) & Abs. ($\uparrow$) & Norm. ($\uparrow$) \\
    \midrule

    Pre-trained (Zero-shot) & 47.72 & - & 55.70 & - \\
    \midrule
    
    Non-linear FT \citep{ilharco2023editing} & 71.25 & 76.94 & 65.04 & 87.98 \\
    \midrule

    TIES-Merging \citep{yadav2023ties} & 74.79 & 82.84 & 62.53 & 94.83 \\
    \midrule

    Task-wise AdaMerging \citep{AdaMerging_ICLR_2024} & 73.39 & 79.02 & 66.19 & 89.86 \\
    Layer-wise AdaMerging \citep{AdaMerging_ICLR_2024} & 77.06 & 82.98 & \ul{66.61} & 89.86 \\
    \midrule

    \textbf{\shortname (Ours)} & 79.67 & 90.73 & 65.04 & 97.22 \\
    \textbf{\shortname + TIES-Merging} & 78.15 & 89.10 & 54.54 & 85.42 \\
    \textbf{\shortname + Task-wise AdaMerging} & \ul{79.73} & \ul{90.84} & 66.47 & \ul{99.21} \\
    \textbf{\shortname + Layer-wise AdaMerging} & \tb{80.25} & \tb{91.40} & \tb{66.76} & \tb{99.63} \\
    
    \bottomrule
    \end{tabular}%
    }
    %\vspace{1.5mm}
    \caption{\textbf{\shortname on different model merging schemes.} Average absolute accuracies (\%) and normalized accuracies (\%) of CLIP ViT-B/32 and T5-Small pre-trained models edited by adding task vectors on each of the downstream tasks. We normalize performance of each method by their single-task accuracy. %For each of the families of methods, 
    \textbf{Bold} indicates the best results. \underline{Underline} the second best.}
    \label{tab:additional_merging}
    \vspace{-3mm}
\end{table}

\subsection{Combining \shortname With Other Model Merging Schemes}\label{sec:talos+modelmerging}

We extend Table \ref{tab:task_addition} in Table \ref{tab:additional_merging} by testing our \shortname in combination with other merging schemes (TIES-Merging \citet{yadav2023ties} and AdaMerging \citet{AdaMerging_ICLR_2024}). Specifically, for TIES-Merging we skip the sparsification part, as the task vectors obtained by \shortname are already sparse. Regarding AdaMerging, we test both Task-wise AdaMerging and Layer-wise AdaMerging.
As we can see, in both vision and language experiments, applying TIES-Merging to our \shortname is harmful. Seemingly, the signs of task vectors obtained via \shortname play an important role and disrupting them according to some heuristics causes a drop in performance. Regarding AdaMerging, we can see that \shortname has full compatibility with existing methods for automating the selection of optimal merging coefficients, highlighting its versatility. However, by itself \shortname is already robust enough that it doesn’t benefit this much from neither task-wise tunings nor layer-wise tunings.

\end{document}